\documentclass[10pt,journal,compsoc]{IEEEtran}
\ifCLASSOPTIONcompsoc
  \usepackage[nocompress]{cite}
\else
  \usepackage{cite}
\fi
\usepackage{graphicx}
\usepackage{cite}
\usepackage{picinpar}
\usepackage{amsmath}
\usepackage{url}
\usepackage{flushend}
\usepackage{amsfonts}
\usepackage{colortbl}
\usepackage{soul}
\usepackage{multirow}
\usepackage{pifont}
\usepackage{color}
\usepackage{alltt}
\usepackage[hidelinks]{hyperref}
\usepackage{enumerate}
\usepackage{siunitx}
\usepackage{breakurl}
\usepackage{epstopdf}
\usepackage{pbox}
\usepackage[table]{xcolor}
\usepackage{placeins}
\usepackage{dblfloatfix}
\usepackage{bm}
\usepackage{booktabs}
\usepackage{multirow}
\usepackage{tabu}
\usepackage{dblfloatfix}
\usepackage{array}
\usepackage{algorithm,algorithmic}
\usepackage{threeparttable}
\usepackage{booktabs, makecell}
\ifCLASSINFOpdf
\else
\fi
\ifCLASSOPTIONcompsoc
 \usepackage[caption=false,font=footnotesize,labelfont=sf,textfont=sf]{subfig}
\else
 \usepackage[caption=false,font=footnotesize]{subfig}
\fi
\begin{document}
%
\title{Teacher-Explorer-Student Learning: A Novel Learning Method for Open Set Recognition}
%
%
%
%

\author{Jaeyeon~Jang and
        Chang Ouk Kim
\IEEEcompsocitemizethanks{\IEEEcompsocthanksitem The authors are with the Department of Industrial Engineering, Yonsei University, Seoul 03722, Republic of Korea.
\protect\\
E-mail: \{jjy009, kimco\}@yonsei.ac.kr}
}

%
%

\markboth{Journal of \LaTeX\ Class Files,~Vol.~14, No.~8, August~2015}%
{Shell \MakeLowercase{\textit{et al.}}: Bare Demo of IEEEtran.cls for Computer Society Journals}
%



\IEEEtitleabstractindextext{%
\begin{abstract}
If an unknown example that is not seen during training appears, most recognition systems usually produce overgeneralized results and determine that the example belongs to one of the known classes. To address this problem, teacher-explorer-student (T/E/S) learning, which adopts the concept of open set recognition (OSR) that aims to reject unknown samples while minimizing the loss of classification performance on known samples, is proposed in this study. In this novel learning method, overgeneralization of deep learning classifiers is significantly reduced by exploring various possibilities of unknowns. Here, the teacher network extracts some hints about unknowns by distilling the pretrained knowledge about knowns and delivers this distilled knowledge to the student. After learning the distilled knowledge, the student network shares the learned information with the explorer network. Then, the explorer network shares its exploration results by generating unknown-like samples and feeding the samples to the student network. By repeating this alternating learning process, the student network experiences a variety of synthetic unknowns, reducing overgeneralization. Extensive experiments were conducted, and the experimental results showed that each component proposed in this paper significantly contributes to the improvement in OSR performance. As a result, the proposed T/E/S learning method outperformed current state-of-the-art methods.
\end{abstract}

}

\maketitle

\IEEEdisplaynontitleabstractindextext

%
\IEEEpeerreviewmaketitle

\IEEEraisesectionheading{\section{Introduction}\label{sec:introduction}}

%
%
%
%
\IEEEPARstart{R}{ecognition} systems have greatly improved due to recent advancements in deep learning \cite{Krizhevsky2012, Simonyan2015, He2016}. However, there are still many challenges to solve in order to apply deep learning techniques to real-world problems. One of the main challenges is that most recognition systems have been designed under closed world assumptions in which all categories are known a priori. However, samples that are unknown in the training phase can be fed into the systems during the testing phase. When an unknown sample appears, traditional recognition systems wrongly identify the sample as belonging to one of the classes learned during training. To handle this problem, the concept of \textit{open set recognition} (OSR), which aims to correctly classify samples of known classes while rejecting unknown samples, has been proposed \cite{Scheirer2013}. In addition, OSR has been introduced in many application areas, including autonomous driving \cite{Wong2019, Pires2020}, network intrusion detection \cite{Henrydoss2017, Cruz2017}, defect classification \cite{S, Jung2020}, and social media forensics \cite{Rocha2017}.

Most existing discriminative models, including deep neural networks (DNNs), suffer from the problem of \textit{overgeneralization} in open set scenarios \cite{Spigler2019}. Here, the overgeneralization problem refers to the situation in which a discriminative model determines with high confidence that unknown samples belong to known classes. Accordingly, many studies have tried to mitigate the overgeneralization problem of OSR. For instance, post recognition score analysis methods were applied in \cite{Bendale2016, Shu2017, Jang2020} to reduce the overgeneralization problem of the output scores of a DNN. Reconstructive and generative networks have also been utilized to calibrate discriminative DNN output scores for supplementary purposes \cite{Yoshihashi2019, Ge2017, Neal2018}. Recently, some researchers proposed two-stage methods that implement an unknown detection task and then a closed set classification task only on samples determined as known; these methods are based on the intuition that minimizing the misclassification of unknown samples is the key to high-performance OSR \cite{Oza2019a, Oza2019, Sun2020}.

\begin{figure}[h]
\centering
\subfloat[]{\includegraphics[width=0.5\linewidth]{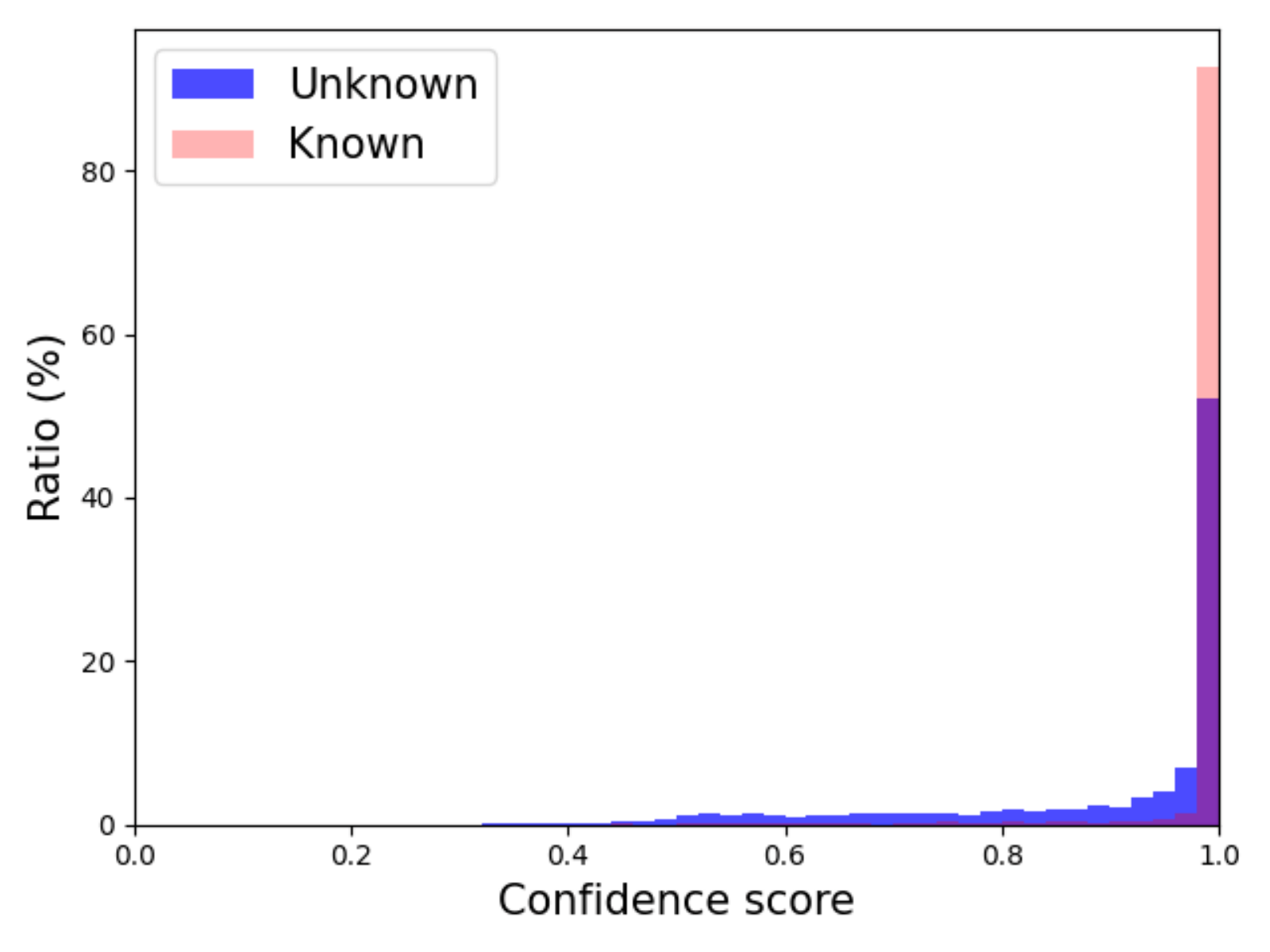}%
\label{fig_1a}}
\subfloat[]{\includegraphics[width=0.5\linewidth]{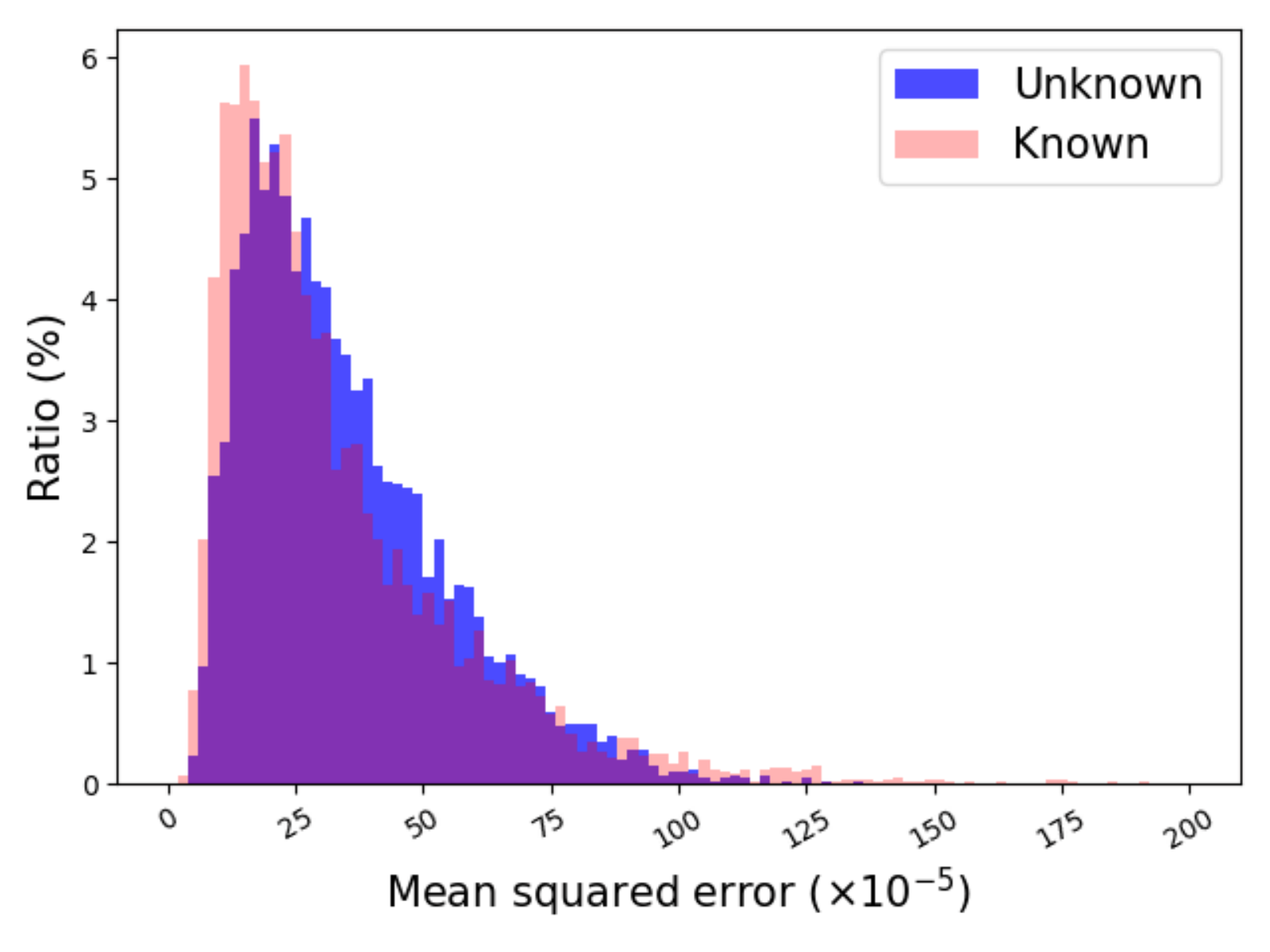}%
\label{fig_1b}}
\par
\subfloat[]{\includegraphics[width=\linewidth]{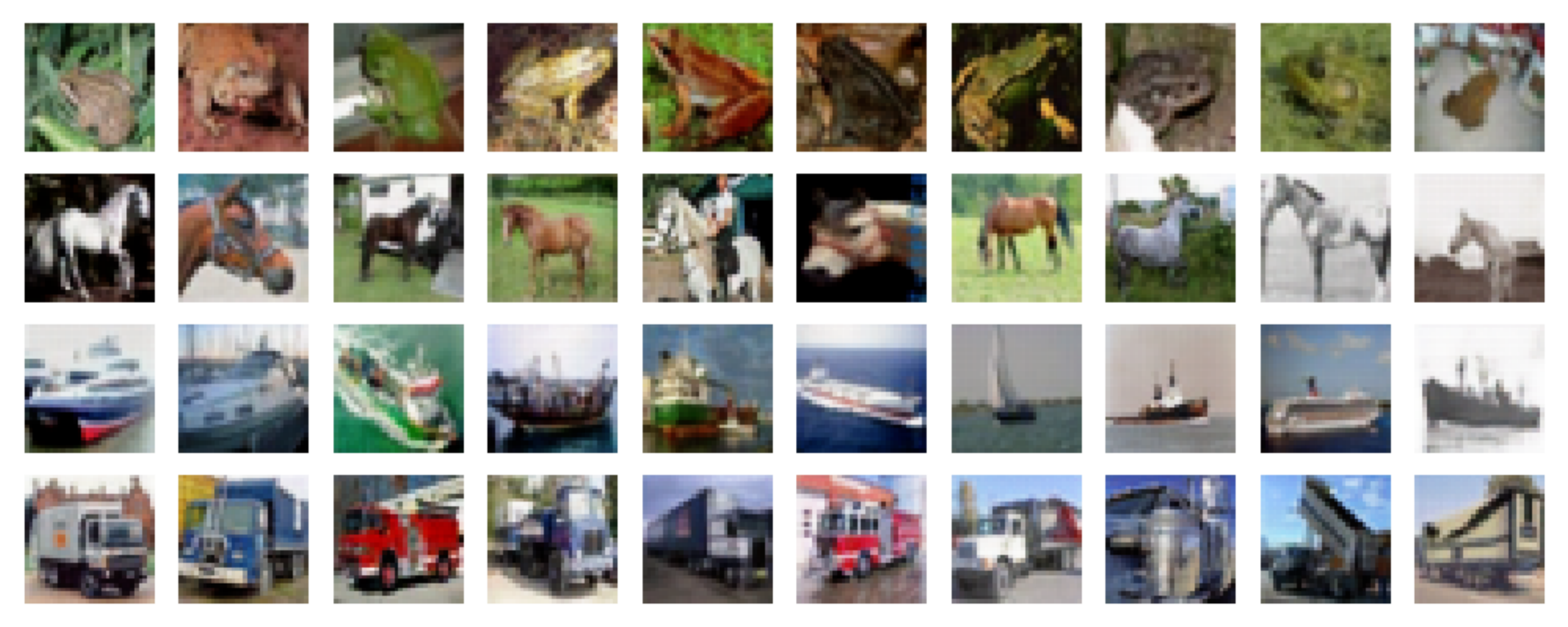}%
\label{fig_1c}}
\caption{Overgeneralization of CNNs and autoencoders. The CIFAR-10 dataset was randomly partitioned into six known classes and four unknown classes, and only known classes’ samples were trained. For the CNN, the maximum score among the known classes was used as the confidence score. (a) Distribution of the CNN confidence score, (b) reconstruction error distribution of the autoencoder, and (c) images of unknown classes reconstructed by the autoencoder.}
\label{fig_1}
\end{figure}

Despite the performance improvement, many OSR methods are still affected by overgeneralization. This is because learning only given known samples, regardless of the type of model used, has limitations in reducing overgeneralization. Fig. \ref{fig_1} shows that not only convolutional neural networks (CNNs), the most commonly used discriminative model for OSR, but also autoencoders, the most commonly used auxiliary model for OSR, produce overly generalized results for unknowns and leave little distinction between knowns and unknowns.

\begin{figure}[h]\centering
  \includegraphics[width=\linewidth]{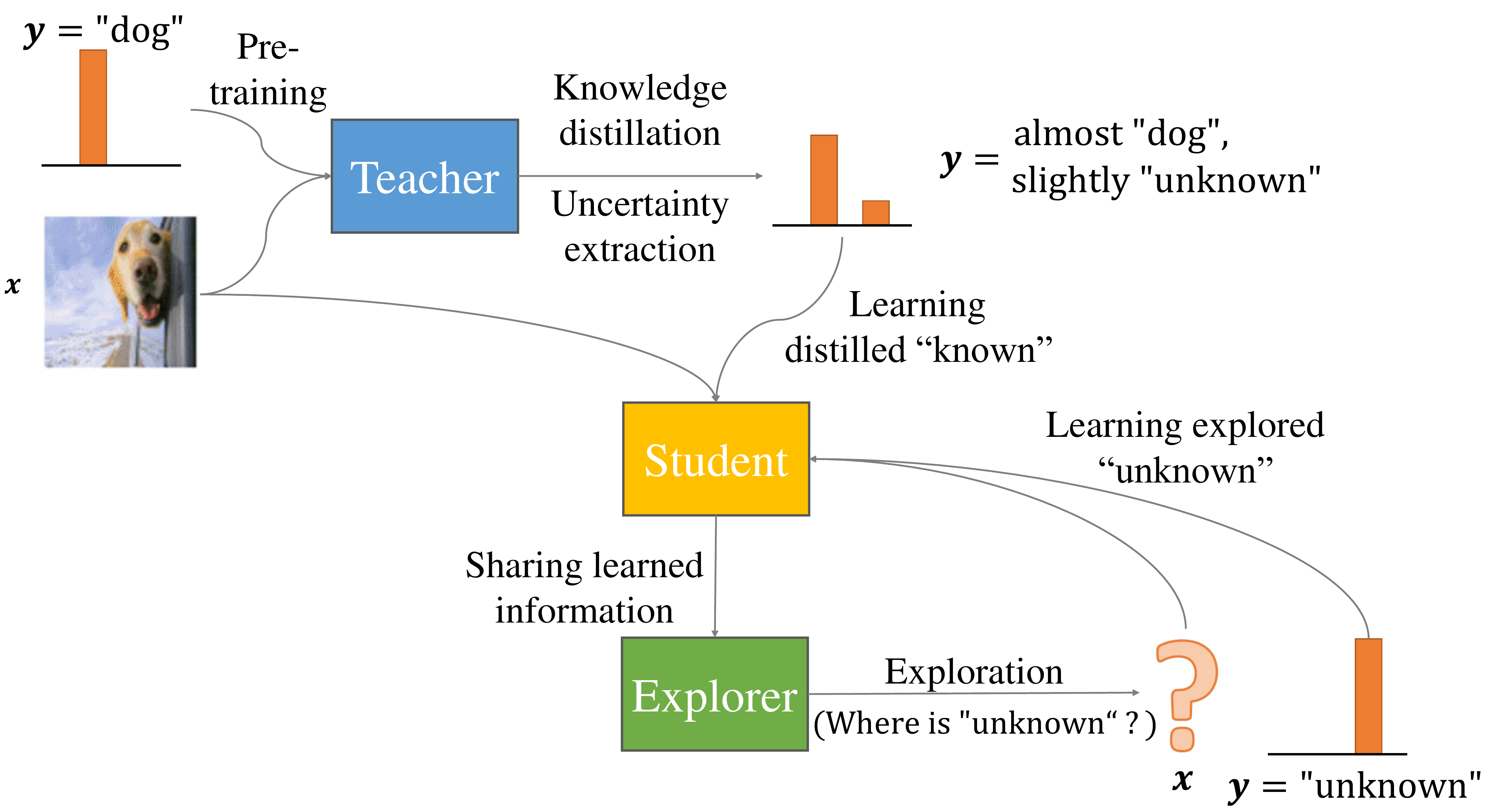}
  \caption{Illustration of teacher-explorer-student learning.}
  \label{fig_2}
\vspace{-10pt}
\end{figure}

Given the infinite diversity of unknowns, the learning of OSR systems must be able to explore various possibilities of unknowns. In this paper, we propose a \textit{teacher-explorer-student} (T/E/S) learning method, as shown in Fig. \ref{fig_2}. Let us assume that there is a student network without any knowledge and a teacher network that has been pretrained and is assumed to have knowledge of the known classes. This teacher network is also assumed to consider the possibilities of unknowns. Then, to deliver the possibilities of unknowns, the teacher must teach not only the original class information of a given example but also the uncertainty that may be inherent in the example. Here, uncertainty is defined as the possibility of belonging to unknown classes. Thus, the teacher distills the information while extracting uncertainty from the example.

Now, the teacher can provide slight hints about unknowns. However, these hints are not sufficient to learn the various possibilities of unknowns. Therefore, we introduce an explorer, a generative adversarial network (GAN), that explores to produce unknown-like open set examples based on the shared information that the student has learned. Finally, the student gains distilled known and explored unknown knowledge, both of which are used to reduce overgeneralization. By repeating this alternating learning process, the student experiences various possibilities of unknowns. In addition, we apply an architecture in which a set of one-vs-rest networks (OVRNs) follow a CNN feature extractor to enable the student network to establish more sophisticated decision boundaries for OSR \cite{Jang2020}.

Extensive experiments were conducted to evaluate the proposed T/E/S learning model. The experimental results showed that the teacher’s distilled knowledge reduces overgeneralization. In addition, the explorer generates realistic but unknown-like synthetic samples that guide the student network to establish tighter decision boundaries. Accordingly, the proposed method outperformed state-of-the-art methods in terms of OSR performance for various open set scenarios.

\section{Background and Related Works}\label{sec:background}
\subsection{Open Set Recognition}
The OSR problem was formalized in \cite{Scheirer2013} as finding a measurable recognition function that minimizes open set risk consisting of open space risk and empirical risk. Here, open space risk is the relative measure of positively labeled open space, which is far from any known training samples, compared to the overall measure of positively labeled space, while empirical risk represents the loss of classification performance on known samples. In the early days of OSR research, some shallow machine learning models were redesigned to introduce open set risk minimization in modeling. For example, Scheirer \textit{et al.} \cite{Scheirer2013} proposed a one-vs-set machine, a variant of the support vector machine (SVM), which introduces an open set risk minimization term into linear kernel SVM modeling. Similarly, Cevikalp \cite{Cevikalp2017} applied the intuitive idea of training a classwise hyperplane to be as close as possible to the target class samples and as far as possible from the other class samples. Scheirer \textit{et al.} \cite{Scheirer2014} introduced statistical extreme value theory (EVT) \cite{Scheirer2010} to calibrate the decision scores of a radial basis function SVM based on the distribution of extreme scores. In addition, they developed a compact abating probability model based on a one-class SVM to manage open space risk. Zhang and Patel \cite{Zhang2017} proposed a sparse representation-based OSR method based on their observation that discriminative information is hidden in the tail of matched and nonmatched reconstruction error distributions.

Over the past few years, deep learning techniques have led to the advancement of OSR systems. Most methods in this category have focused on mitigating overgeneralization of general discriminative DNNs that usually generate excessive open space \cite{Spigler2019}. The first deep model introduced for OSR was OpenMax, which models a class-specific representation distribution in the penultimate layer of a CNN and computes the regularized confidence score by applying an EVT-based calibration strategy \cite{Bendale2016}. Shu \textit{et al.} \cite{Shu2017} replaced a softmax layer with a sigmoid layer, whose output nodes make their own class-specific determinations. They additionally applied Gaussian fitting to obtain class-specific reject/accept thresholds that tighten the decision boundaries. Jang and Kim \cite{Jang2020} showed that the architecture in which a set of OVRNs follows a CNN feature extractor enables more discriminative feature learning for OSR. In addition, collective decisions of OVRNs were used to establish more sophisticated decision boundaries that reduce redundant open space.

Some researchers have adopted reconstructive or generative models to calibrate the confidence score of discriminative DNNs. For instance, Yoshihashi \textit{et al.} \cite{Yoshihashi2019} proposed a deep hierarchical reconstruction network (DHRNet) that combines classification and reconstruction networks. They expanded the OpenMax model by additionally utilizing the hierarchical latent representations of DHRNet. Ge \textit{et al.} \cite{Ge2017} further enhanced the OpenMax model by utilizing the synthetic samples generated by a conditional GAN. Neal \textit{et al.} \cite{Neal2018} proposed an encoder-decoder GAN that generates counterfactual samples and retrained a pretrained CNN to classify the generated samples as unknown samples. However, the synthetic samples produced by the two GAN-based methods are limited to only a small portion of the open space \cite{Geng2020}.

Recently, two-stage methods that sequentially implement unknown detection and closed set classification tasks have been at the forefront of advancement. Oza and Patel \cite{Oza2019a} proposed a network configuration in which a decoder and a classifier follow a shared encoder for reconstruction and classification. They model the tail of the reconstruction error distribution with EVT to compute the unknown detection score. Finally, the classifier assigns one class among the known classes for the samples determined as known samples. In a subsequent study \cite{Oza2019}, they extended the decoder into the class-conditioned decoder and defined their model as C2AE. Sun \textit{et al.} \cite{Sun2020} proposed a conditional Gaussian distribution learning (CGDL) method that generates a class-conditional posterior distribution in the latent space using a variational autoencoder, which follows classwise multivariate Gaussian models. The learned features are fed into two models: an unknown detector and a closed set classifier. The unknown detector identifies unknowns based on the set of classwise Gaussian cumulative probabilities and the reconstruction errors of the variational autoencoder.

\subsection{One-vs-rest Networks}
The softmax function is the de facto standard activation used for multiclass classification; it measures the relative likelihood of a known class compared to the other known classes. Due to this characteristic, when an unknown sample is fed, a network with a softmax output layer is trained to choose the best matching class instead of rejecting the sample \cite{Boult2019}. That is, a network with softmax is at high risk of giving a high confidence score to unknowns by selecting the most similar class among all known classes. On the other hand, if sigmoid activation is applied to the output layer, each sigmoid output is not conditioned on the other outputs. Rather, each sigmoid output is trained to discriminate a dissimilar example from the matched examples, allowing all the classes’ output nodes to independently reject unknown examples. Thus, by combining multiple class-specific determinations into the collective decision, more sophisticated decision boundaries for rejection can be established. In addition, the overgeneralization problem is further reduced by applying a set of OVRNs as the output layer instead of a single sigmoid layer \cite{Jang2020}. Thus, we apply a structure in which OVRNs follow a CNN feature extractor to the student network.

\subsection{Teacher-Student Learning}
In this paper, teacher-student (T/S) learning \cite{Romero2015, Hinton2015} is extended for OSR. Thus, in this section, we briefly introduce the original concept of T/S learning. Recent top performing DNNs usually involve very wide and deep structures with numerous parameters. T/S learning, often called knowledge distillation, was proposed to reduce the computational burden of inference caused by the heavy structure.

In original T/S learning, the knowledge of a heavy teacher network is transferred into a relatively light student network. The student network is penalized to learn a softened version of the teacher’s output. Learning this soft target guides the student to capture the finer structure learned by the teacher \cite{Romero2015}. Generally, neural networks produce posterior class probabilities $q_y=\frac{\text{exp}(l_y)}{\sum_{j\in \mathcal{Y}}\text{exp}(l_j)}, \forall y\in \mathcal{Y}$, with softmax activation, where $l_y$ is the logit of a class $y$ and $\mathcal{Y}$ is the set of known classes. To produce the soft targets, the class probabilities are scaled by temperature $\tau$ as follows:
\begin{gather} 
q^\tau_y=\frac{\text{exp}(l_y/\tau)}{\sum_{j\in \mathcal{Y}}\text{exp}(l_j/\tau)}, \forall y\in \mathcal{Y}. \label{eq1}
\end{gather}

In the T/S learning, additional semantic information is provided to the student network by increasing the probabilities of non-target classes. The interest thing is that the student can recognize samples of unseen classes by only learning softened probabilities of the seen classes’ examples, if the teacher has the knowledge about the unseen classes \cite{Hinton2015}. This is because the teacher gives the seen examples a small possibility of belonging to the unseen classes and the student can infer the unseen classes with that small possibility. Without loss of generality, the student network can recognize unknowns, if the teacher network can discover the uncertainties inherent in known samples.

\section{Proposed Method}\label{sec:method}
\begin{figure*}[h]\centering
  \includegraphics[width=0.8\linewidth]{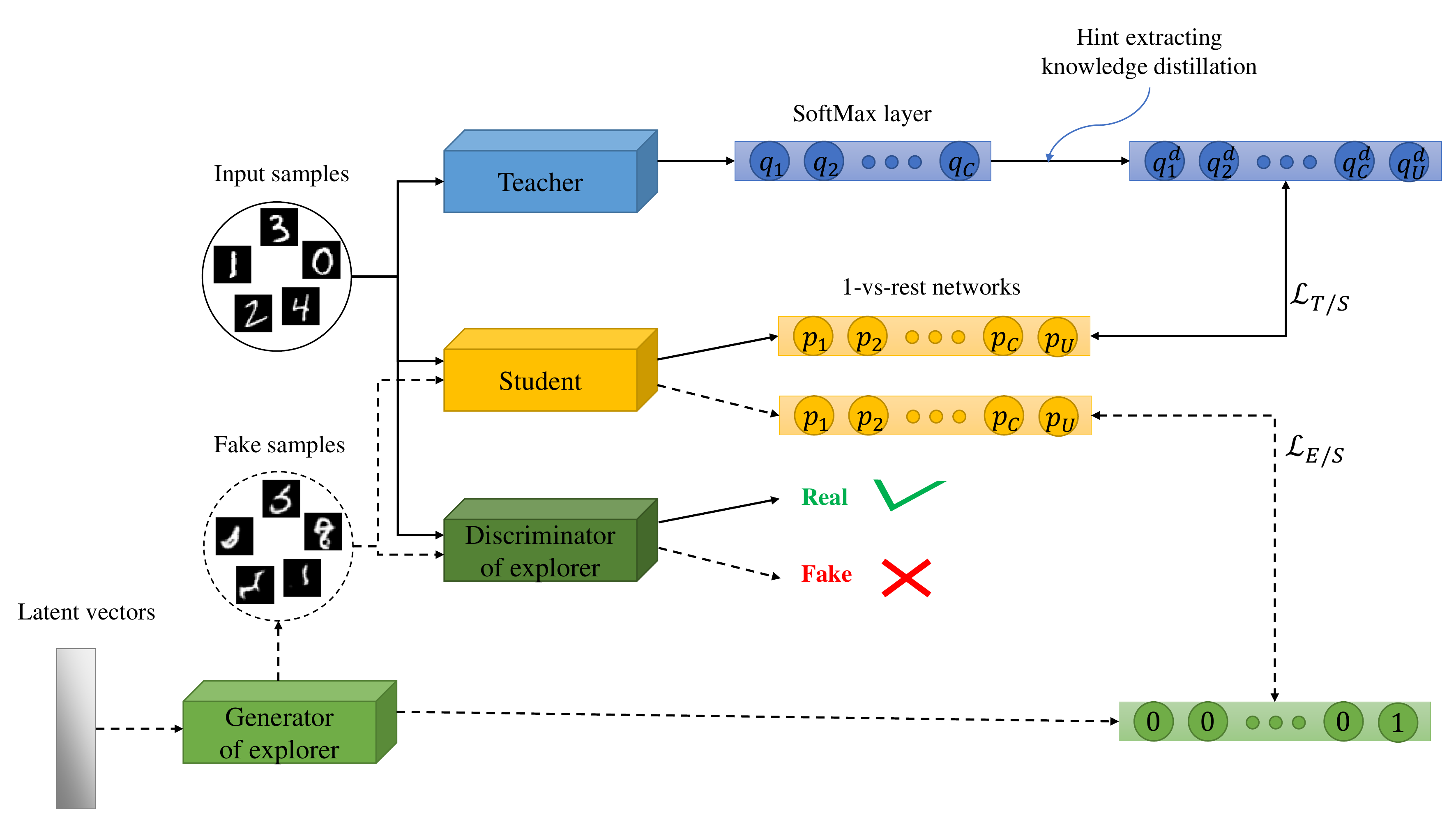}
  \caption{Overview of teacher-explorer-student learning. Single arrows denote the direction of the information flow while double arrows connect two elements of the loss function. Solid lines and dashed lines are used to indicate the flow of real samples and fake samples, respectively.}
  \label{fig_3}
\vspace{-10pt}
\end{figure*}

Fig. \ref{fig_3} shows an overview of the proposed T/E/S learning method. First, the teacher network is pretrained to provide $\bm{q}$, posterior probabilities of known classes. Next, the probabilities are calibrated to assign softened probabilities for the known classes and hints for $U$ that represents all unknown classes. For this calibration, a novel hint extracting knowledge distillation (HE-KD) method is suggested. Intuitively, the student network can recognize an unknown sample well after learning the sufficient and diverse possibilities of unknowns. However, the HE-KD method only gives a small hint about unknowns. To tackle this problem, the explorer network that explore open space and generate unknown-like examples are proposed. Here, the role of the explorer is to support the student by discovering new open set examples based on the current student’s knowledge about $U$. Thus, the student and the explorer are trained together alternately.

Through T/E/S learning, the student learns not only the information about “known” distilled by the teacher but also the information about “unknown” explored by the explorer. In every iteration, real samples and generated samples are fed into the student network. The student network is trained to produce a soft probability vector $\bm{q}^d$ distilled by HE-KD for known training samples by minimizing $\mathcal{L}_{T/S}$, which is the loss between $\bm{p}$ and $\bm{q}^d$, where $\bm{p}$ is the output vector of the student and $\bm{q}^d$ is the distilled probability vector. In addition, for fake samples, the student network is trained to minimize $\mathcal{L}_{E/S}$, which is the loss between $\bm{p}$ and the hard label for $U$.

\subsection{Teacher Network: Hint Extracting Knowledge Distillation}
Let $\bm{x}_i\in \mathcal{X}$ be an input sample from any of the known classes and $t_i\in \mathcal{Y}$ be its corresponding ground-truth label, where $\mathcal{X}$ is the input space. Then, the teacher network ($T$) with parameters $\bm{\theta}_T$ is first trained to minimize the following categorical cross-entropy loss:
\begin{gather} 
\mathcal{L}_T(\bm{\theta}_T)=-\frac{1}{N}\sum_{i=1}^N\sum_{y \in \mathcal{Y}}\mathbb{I}(t_i=y)\text{log}q_{iy}, \label{eq2}
\end{gather}
where $N$ is the batch size, $\mathbb{I}$ is the indicator function, and $q_{iy}$ is the posterior probability of sample $\bm{x}_i$ for class $y$.

After the teacher is trained, the teacher extracts uncertainty from the training samples. The teacher considers a training sample to be more uncertain if the sample has a lower probability for the target class. However, the trained teacher provides a very high target class score for most training samples, leaving no difference among the samples. Thus, we produce a scaled probability vector $\bm{q}_i^\tau$ for $\bm{x}_i$ by applying the temperature scaling provided in \eqref{eq1}.

HE-KD regenerates the distilled target class probability $q_{it_i}^d$ and the uncertainty $q_{iU}^d  (=1-q_{it_i}^d)$ based on the quantity of $q_{it_i}^\tau$. The other elements of $\bm{q}_i^d$ are set to zero. Let the training data be split into $\mathcal{D}_c$ and $\mathcal{D}_m$, which are the set of examples correctly classified by the teacher and the set of misclassified examples, respectively. Let $S_{\mathcal{D}_c}$ be the set of scaled target probabilities $q_{it_i}^\tau$ for $\bm{x}_i\in \mathcal{D}_c$. Then,  $q_{it_i}^d$ is computed as follows:
\begin{gather} 
q_{it_i}^d= \begin{cases}
    q_{\text{min}}^d+(1-q_{\text{min}}^d)N(q_{it_i}^\tau \mid S_{\mathcal{D}_c}) & \text{if } \bm{x}_i \in \mathcal{D}_c\\
    q_{\text{min}}^d & \text{otherwise}
  \end{cases}, \label{eq3}
\end{gather}
where $q_{\text{min}}^d$ is the minimum distilled probability for the target class and $N(q_{it_i}^\tau \mid S_{\mathcal{D}_c})=\frac{q_{it_i}^\tau-\text{min}(S_{\mathcal{D}_c})}{\text{max}(S_{\mathcal{D}_c})-\text{min}(S_{\mathcal{D}_c})}$. Here, $q_{\text{min}}^d$ is a parameter given to prevent the student network from learning too small probability for the target class and losing its discriminative capability.

\subsection{Explorer Networks: Open Set Example Generation}
\begin{figure}[b]\centering
  \includegraphics[width=\linewidth]{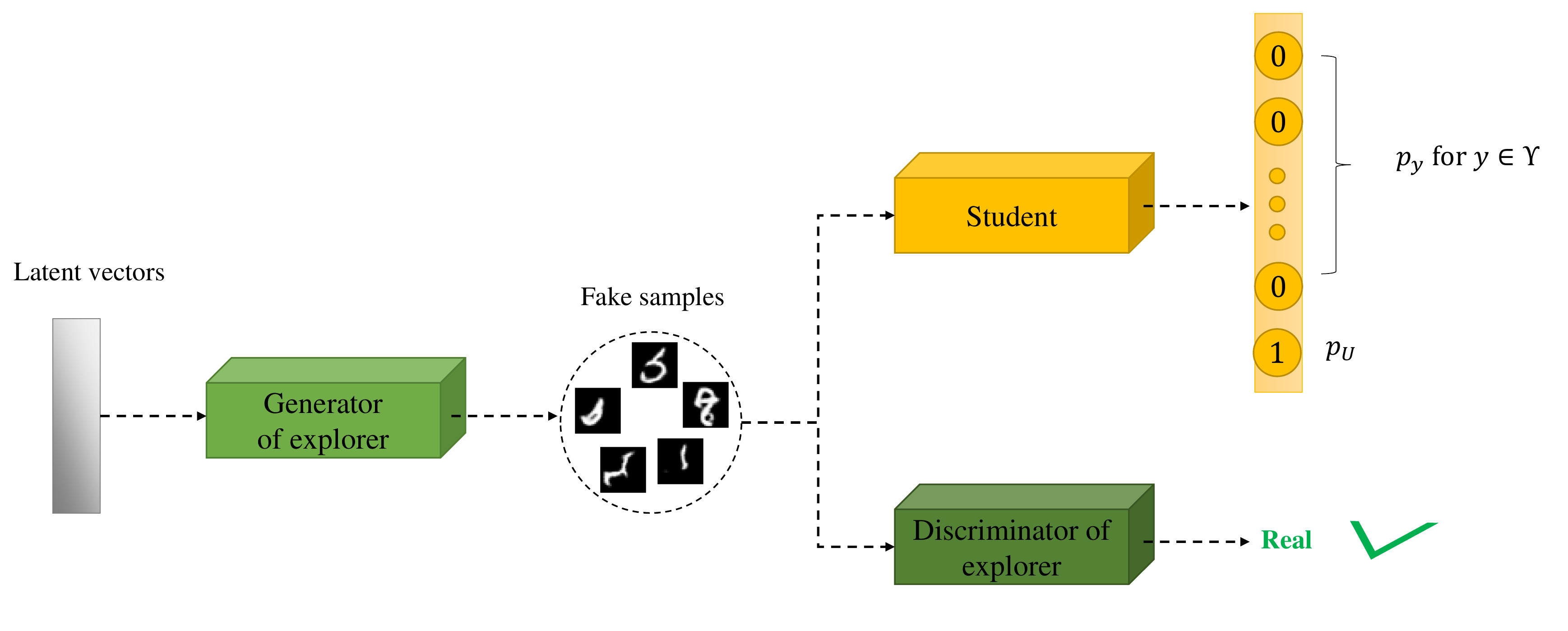}
  \caption{Goal of the explorer’s generator.}
  \label{fig_4}
\vspace{-10pt}
\end{figure}

The explorer networks adopt a general GAN structure containing a generator and a discriminator. In the original form of GAN learning, the goal of the generator is to produce fake samples that deceive the discriminator into identifying the fake samples as real. In addition to this original goal, the generator of the explorer is trained to produce fake samples in open space, which the student determines to be unknown samples, as shown in Fig. \ref{fig_4}. Let $S$, $G$, and $D$ be the student, the generator, and the discriminator, respectively. Let the latent noise vector $\bm{z}$ follow a prior distribution $p_{\text{pri}}(\bm{z})$. Then, the objective function of the generator is as follows:
\begin{gather} 
    \min_{\bm{\theta}_G}\mathbb{E}_{\bm{z}\sim p_{\text{pri}}(\bm{z})}[\text{log}(1-D(G(\bm{z})))+\lambda\mathcal{L}_{BCE}(\bm{y}_U,S(G(\bm{z})))], \label{eq4}
\end{gather}
where $\bm{\theta}_G$ is the generator’s parameter set, $\mathcal{L}_{BCE}(\cdotp,\cdotp)$ is the binary cross entropy, $\bm{y}_U=[0,0,\cdots,1]^\top$ is the hard label of an unknown sample, and $\lambda$ is a balancing parameter.

The discriminator of the explorer is trained to discriminate real and fake samples by updating $\bm{\theta}_D$, the discriminator’s parameter set, based on \eqref{eq5}. Because the discriminator learns the synthetic samples as fake and the training samples as real, whenever the synthetic examples are generated, the generator consistently tries to produce new realistic open set examples at every iteration. The objective function of this alternating training can be expressed as \eqref{eq6}.
\begin{gather} 
    \max_{\bm{\theta}_D}\mathbb{E}_{\bm{x}\sim \mathcal{X}}[\text{log}(D(\bm{x}))]+\mathbb{E}_{\bm{z}\sim p_{\text{pri}}(\bm{z})}[\text{log}(1-D(G(\bm{z})))]. \label{eq5}
\end{gather}
\begin{equation}\label{eq6}
\begin{aligned}
    \min_{\bm{\theta}_G}\max_{\bm{\theta}_D}\mathbb{E}_{\bm{x}\sim \mathcal{X}}[\text{log}(D(\bm{x}))]&+\mathbb{E}_{\bm{z}\sim p_{\text{pri}}(\bm{z})}[\text{log}(1-D(G(\bm{z})))\\&+\lambda\mathcal{L}_{BCE}(\bm{y}_U,S(G(\bm{z})))].
\end{aligned}
\end{equation}

\subsection{Student Network: Learning Known and Unknown-like Samples}
In T/E/S learning, the student network learns real known samples and open set samples generated by the explorer. For a real sample $\bm{x}_i$, the student is trained to predict $\bm{q}_i^d$ based on the following binary cross-entropy loss function:
\begin{equation}\label{eq7}
\begin{aligned}
\mathcal{L}_{T/S}(\bm{\theta}_S)=&-\frac{1}{N}\sum_{i=1}^N\sum_{y \in \mathcal{Y}\cup \{U\}}[q_{iy}^d\text{log}S(\bm{x}_i)_y\\&+(1-q_{iy}^d)(1-\text{log}S(\bm{x}_i)_y)],
\end{aligned}
\end{equation}
where $S(\bm{x}_i)_y$ is $p_{iy}$, the student output of $\bm{x}_i$ for class $y$, and $\bm{\theta}_S$ denotes the student’s parameter set.

The student also learns fake samples $\tilde{\bm{x}}_k=G(\bm{z}_k), \bm{z}_k\sim p_{\text{pri}}(\bm{z})$. However, it is dangerous to let the student network train all $\tilde{\bm{x}}_k$ with a hard label $\bm{y}_U$. This is because in the competitive learning employed by the explorer, the generator sometimes produces known-like samples. Training the known-like samples as unknown samples can decrease the closed set classification performance. Thus, only unknown-like samples, which we call active unknown samples, are used in the training of the student network. The active unknown samples are selected by the indicator function $\mathcal{A}$, given as follows:
\begin{gather} 
\mathcal{A}(\tilde{\bm{x}}_k)= \begin{cases}
    1 & \text{if } \max_{y \in \mathcal{Y}}S(\tilde{\bm{x}}_k)_y<1-q_{\text{min}}^d\\
    0 & \text{otherwise}
  \end{cases}. \label{eq8}
\end{gather}
Here, the criterion $1-q_{\text{min}}^d$ is the maximum uncertainty for the known training samples. When the generator produces the same number of fake samples as the input batch of real samples, the student is trained to minimize (9).
\begin{equation}\label{eq9}
\begin{aligned}
&\mathcal{L}_{S}(\bm{\theta}_S)=\\&-\frac{1}{N}\sum_{i=1}^N\sum_{y \in \mathcal{Y}\cup \{U\}}[q_{iy}^d\text{log}S(\bm{x}_i)_y+(1-q_{iy}^d)(1-\text{log}S(\bm{x}_i)_y)]\\&-\frac{1}{N}\sum_{k=1}^N[\mathcal{A}(\tilde{\bm{x}}_k)(\text{log}S(\tilde{\bm{x}}_k)_U+\sum_{y \in \mathcal{Y}}(1-\text{log}S(\tilde{\bm{x}}_k)_y))]
\end{aligned}
\end{equation}

The student network is jointly trained with the explorer while sharing learned information. The joint training is summarized in Algorithm \ref{alg_1}.

\begin{algorithm} 
 \caption{Pseudo code for joint training}
\label{alg_1}
 \begin{algorithmic}[1]
 \renewcommand{\algorithmicrequire}{\textbf{Require:} }
 \REQUIRE Pretrained teacher network $T$,  student network $S$, generator $G$, discriminator $D$, training dataset $\mathfrak{D}$;
 \\ 
  \STATE Initialize empty distilled probability vector set $\bm{Q}^d$;
  \STATE Set batch size $N$;
  \STATE Compute $L$, the number of iterations per epoch;
 \\
  \FOR {all $\bm{x}_i \in \mathfrak{D}$}
  \STATE Compute $\bm{q}_i^d$ using \eqref{eq1} and \eqref{eq3};
  \STATE $\bm{Q}^d \gets \bm{Q}^d\cup \{\bm{q}_i^d\}$;
  \ENDFOR
 \WHILE{$\bm{\theta}_S$ has not converged}
  \FOR {$l = 1, 2, \cdots, L$}
  \STATE Initialize empty fake sample set $B_{\tilde{\mathcal{X}}}$;
  \STATE Sample $B_{\mathcal{X}}=\{\bm{x}_{(1)}, \bm{x}_{(2)}, \cdots, \bm{x}_{(N)}\}$;
  \STATE Sample $B_{\text{pri}}=\{\bm{z}_{(1)}, \bm{z}_{(2)}, \cdots, \bm{z}_{(N)}\}$;
  \STATE Update $\bm{\theta}_D$ by feeding $B_{\mathcal{X}}$ and $B_{\text{pri}}$ based on \eqref{eq5};
  \STATE Update $\bm{\theta}_G$ by feeding $B_{\text{pri}}$ based on \eqref{eq4};
  \FOR {all $\bm{z}_k \in B_{\text{pri}}$}
\STATE Generate $\tilde{\bm{x}}_k=G(\bm{z}_k)$;
\STATE $B_{\tilde{\mathcal{X}}} \gets B_{\tilde{\mathcal{X}}}\cup \{\tilde{\bm{x}}_k\}$;
  \ENDFOR
  \STATE Update $\bm{\theta}_S$ by feeding $B_{\mathcal{X}}$ and $B_{\tilde{\mathcal{X}}}$ based on \eqref{eq9}
  \ENDFOR
\ENDWHILE
 \end{algorithmic} 
 \end{algorithm}
\vspace{-10pt}

\subsection{Open Set Recognition Rule}
In this section, we propose a recognition rule based on the collective decisions of OVRNs in the student network. A sample is more likely to belong to the target class when the sample has a high score for the target class output and low scores for the other classes. Furthermore, since nontarget OVRNs usually produce zero probability for a sample, we compute the collective decision score based on the logits of the OVRNs as suggested in \cite{Jang2020}. Let $l_{iy}$ be the logit value of example $\bm{x}_i$ for class $y$. Then, $cds_{iy}$, the collective decision score for class {y}, is computed with the following simple function:
\begin{gather} 
cds_{iy}=l_{iy}-\frac{1}{\vert\mathcal{Y}\vert}\sum_{\substack{t \in \mathcal{Y}\cup \{U\}\\ t\neq y}}l_{it} \; \forall y \in \mathcal{Y}\cup \{U\}. \label{eq10}
\end{gather}

Additionally, the unknown probability score $p_{iU}$ can be used individually to supplement unknown detection because the OVRN of the unknown class $U$ is trained to discriminate between known samples and unknown-like samples. Thus, we propose an OSR rule for both closed set classification and unknown detection as follows:
\newcommand{\argmax}{\arg\!\max}
\begin{gather} 
y^*= \begin{cases}
\argmax_{y \in \mathcal{Y}\cup \{U\}}cds_{iy} & \text{if } cds_{iy} > \epsilon_y^{cds} \text{and }\\ &
p_{iU}<\epsilon^U \text{(optional)}
\\
    U & \text{otherwise}
  \end{cases}, \label{eq11}
\end{gather}
where $\epsilon_y^{cds}$ is the collective decision score threshold for a class $y$ and $\epsilon^U$ is the threshold of the uncertainty. Empirically, it is not recommended to apply the condition $p_{iU}<\epsilon^U$ if it is not expected that there will be many unknown samples during the testing phase.

\begin{figure*}[b]\centering
  \includegraphics[width=0.8\linewidth]{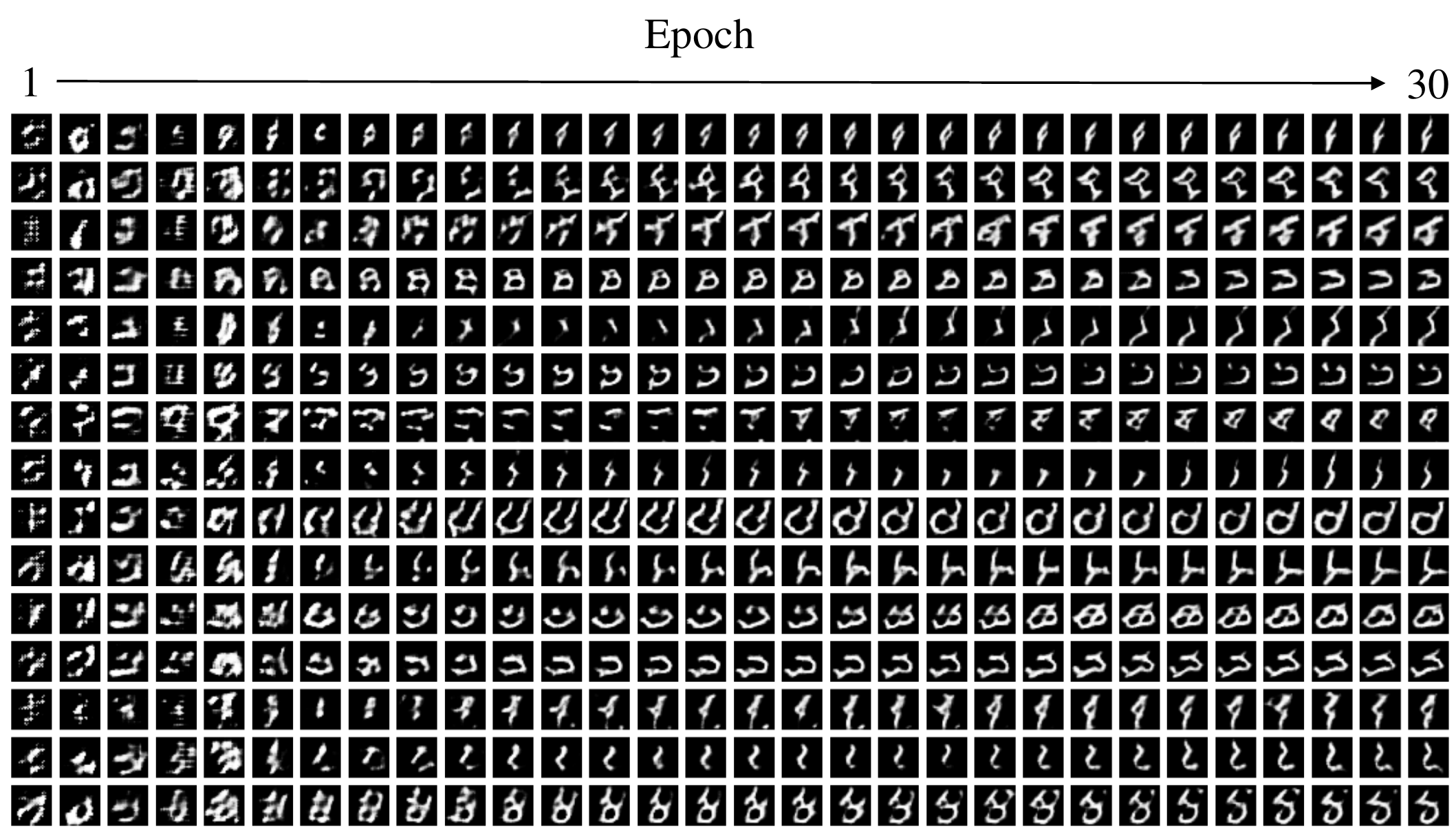}
  \caption{Change in the MNIST open set examples during training.}
  \label{fig_5}
\vspace{-10pt}
\end{figure*}

\section{EXPERIMENTS}
To evaluate the performance of the T/E/S learning method, extensive experiments were conducted on multiple benchmark datasets. First, we analyzed how the generated open set examples affect learning in open set scenarios. Then, various ablative experiments were conducted to validate the contribution of each component in the proposed learning method. Finally, we compared the proposed method with state-of-the-art OSR methods.

\subsection{Implementation Details}
We employed the two CNN configurations suggested in \cite{Jang2020}, which added a set of OVRNs to the plain CNN and the redesigned VGGNet defined in \cite{Yoshihashi2019}, for the student networks. For the teacher networks, the original versions suggested in \cite{Yoshihashi2019} were used. Specifically, the plain CNN was used for training the MNIST dataset and the redesigned VGGNet was used for training the other datasets. Finally, we applied the architectures shown in Table \ref{table_1} for the explorers. We used the Adam optimizer \cite{Kingma2014} with a learning rate of 0.002 for all networks used in the experiments. A class-specific threshold $\epsilon_y^{cds}$ was obtained by ensuring that 95\% of class $y$ training data were recognized as known and classified as class $y$. $\epsilon_U^{cds}$ was set as zero. In addition, $\epsilon_U$ was set to ensure that 95\% of the training data were recognized as known. The minimum distilled probability $q_{min}^d$ in \eqref{eq3} and \eqref{eq8} was empirically set to 0.7. To set temperature $\tau$ in \eqref{eq1} and balancing parameter $\lambda$ in \eqref{eq4}, we applied a cross-class validation framework \cite{Bendale2015}, which measures the performance on the validation set while leaving out a randomly selected subset of known classes as “unknown”.

\newcolumntype{C}[1]{>{\centering\arraybackslash}p{#1}}
\begin{table}[h]
\centering
\begin{threeparttable}
\caption{Explorer Network Architectures}
\label{table_1}
\begin{tabular}{C{0.4\linewidth}|C{0.4\linewidth}}
\hline\hline
\multicolumn{2}{c}{MNIST}\\
\hline									
Generator & Discriminator \\
\hline	
Input: 100 & Input: (28, 28, 1)\\
FC($7\times7\times128$) & C(64, 3, 2)\\
R(7, 7, 128) & C(64, 3, 2)\\
TC(128, 4, 2) & FC(1)\\
TC(128, 4, 2) & Output: 1\\
C(1, 7, 1) & \\
Output: (28, 28, 1) & \\ 
\hline\hline
\multicolumn{2}{c}{}\\
\hline\hline
\multicolumn{2}{c}{Others}\\
\hline									
Generator & Discriminator \\
\hline	
Input: 100 & Input: (32, 32, 3)\\
FC($4\times4\times256$) & C(64, 3, 2)\\
R(4, 4, 256) & C(128, 3, 2)\\
TC(128, 4, 2) & C(128, 3, 2)\\
TC(128, 4, 2) & C(256, 3, 2)\\
TC(128, 4, 2) & FC(1)\\
C(3, 3, 1) & Output: 1\\
Output: (32, 32, 3) & \\ 
\hline\hline
\end{tabular}
    \begin{tablenotes}
      \small
      \item \textit{FC($x$) is a fully connected layer with $x$ nodes. R is a reshape layer. C($x, y, z$) and TC($x, y, z$) are a convolutional layer and a transposed convolutional layer with $x$ filters, a $y\times y$ kernel, and a stride=$z$. Sigmoid activation is used in the output layer and leaky ReLU is used in the other layers.}
    \end{tablenotes}
\end{threeparttable}
\vspace{-10pt}
\end{table}

\subsection{Effects of Open Set Example Generation}
In this section, the effects of the explorer are addressed by analyzing the open set examples generated by the explorer. First, we show the open set examples generated by the explorer. We then analyze the change in the distribution of the active unknown examples, the open set examples participating in training, during the training phase. A 2-dimensional toy dataset was used for this experiment. The goal of these two analyses is to reveal the role of the generated open set examples in T/E/S learning.  Finally, we analyze how unknown samples and open set samples are recognized by a trained student network. 

\begin{figure*}[b]
\begin{minipage}{0.49\textwidth}\centering
  \includegraphics[width=0.7\linewidth]{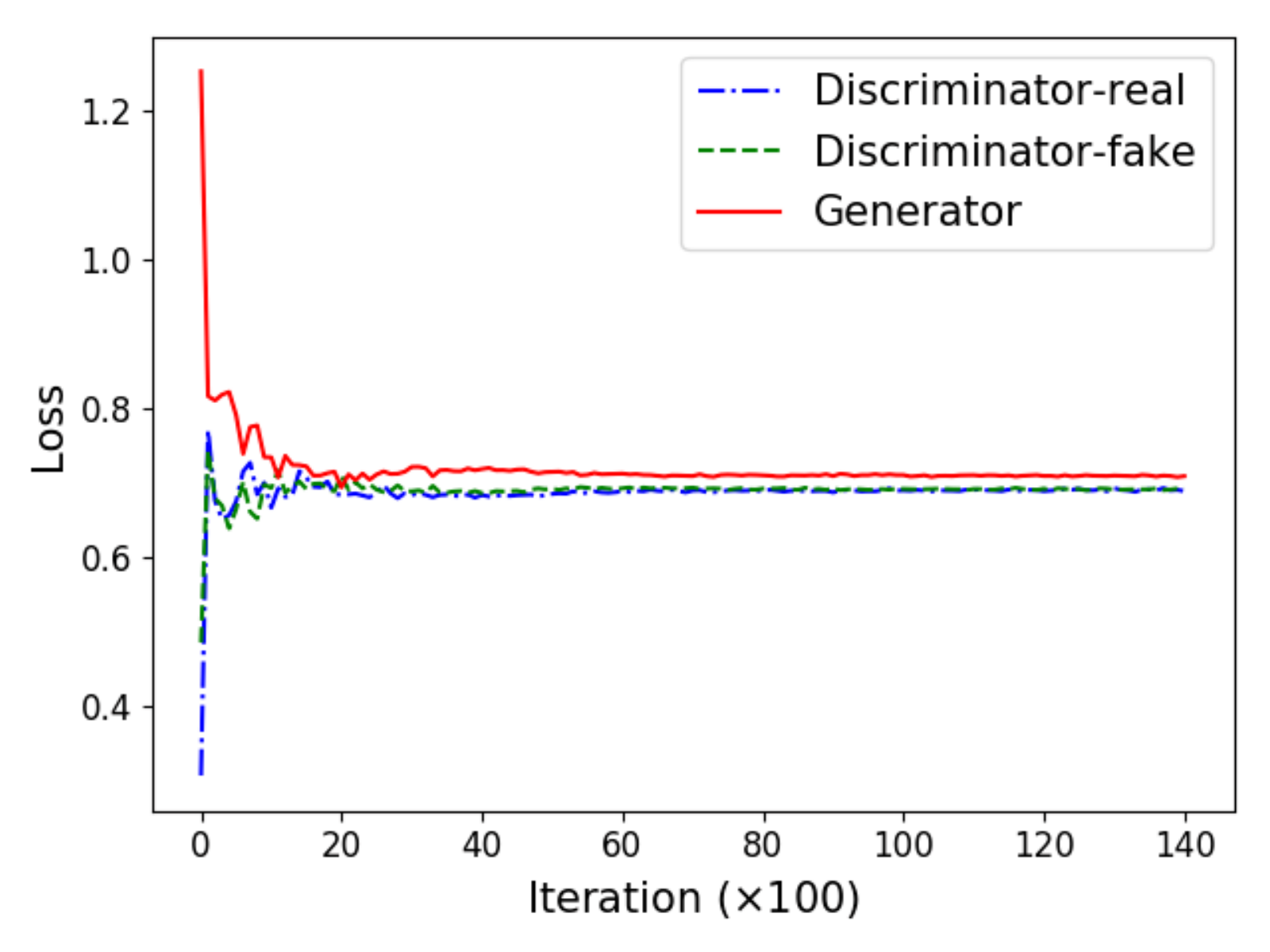}
  \caption{Loss of the discriminator and generator during MNIST training. For the generator, the loss of deceiving the discriminator was measured.}
  \label{fig_6}
\end{minipage}
\hspace{0.01\textwidth}
\begin{minipage}{0.49\textwidth}\centering
  \includegraphics[width=0.7\linewidth]{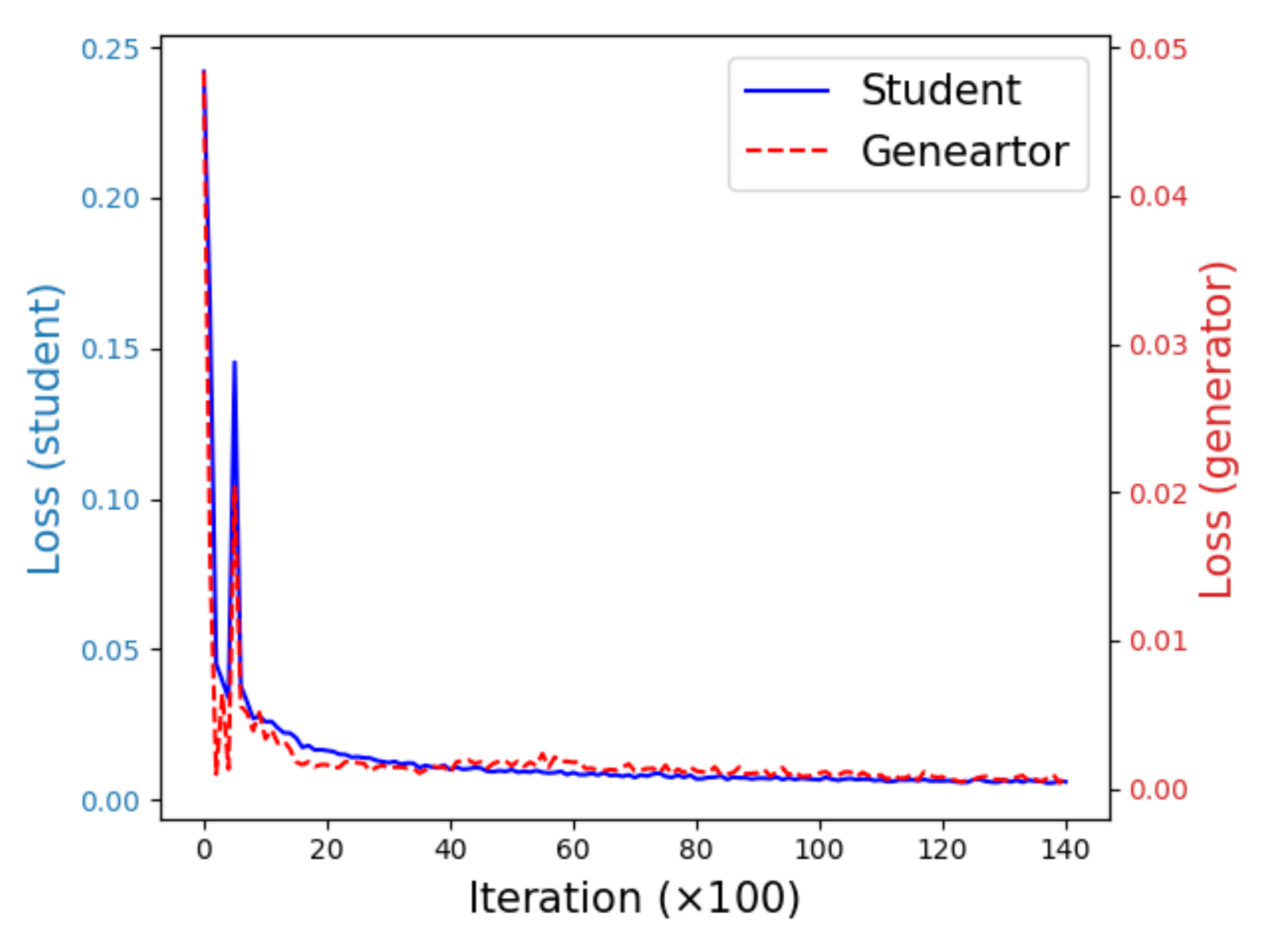}
  \caption{Loss of the student network during MNIST training. In the graph, the student and generator represent the loss for real training samples and active unknowns, respectively.}
  \label{fig_7}
\end{minipage}
\begin{minipage}{\textwidth}
\centering
\subfloat[Epoch 19]{\includegraphics[width=0.3\linewidth]{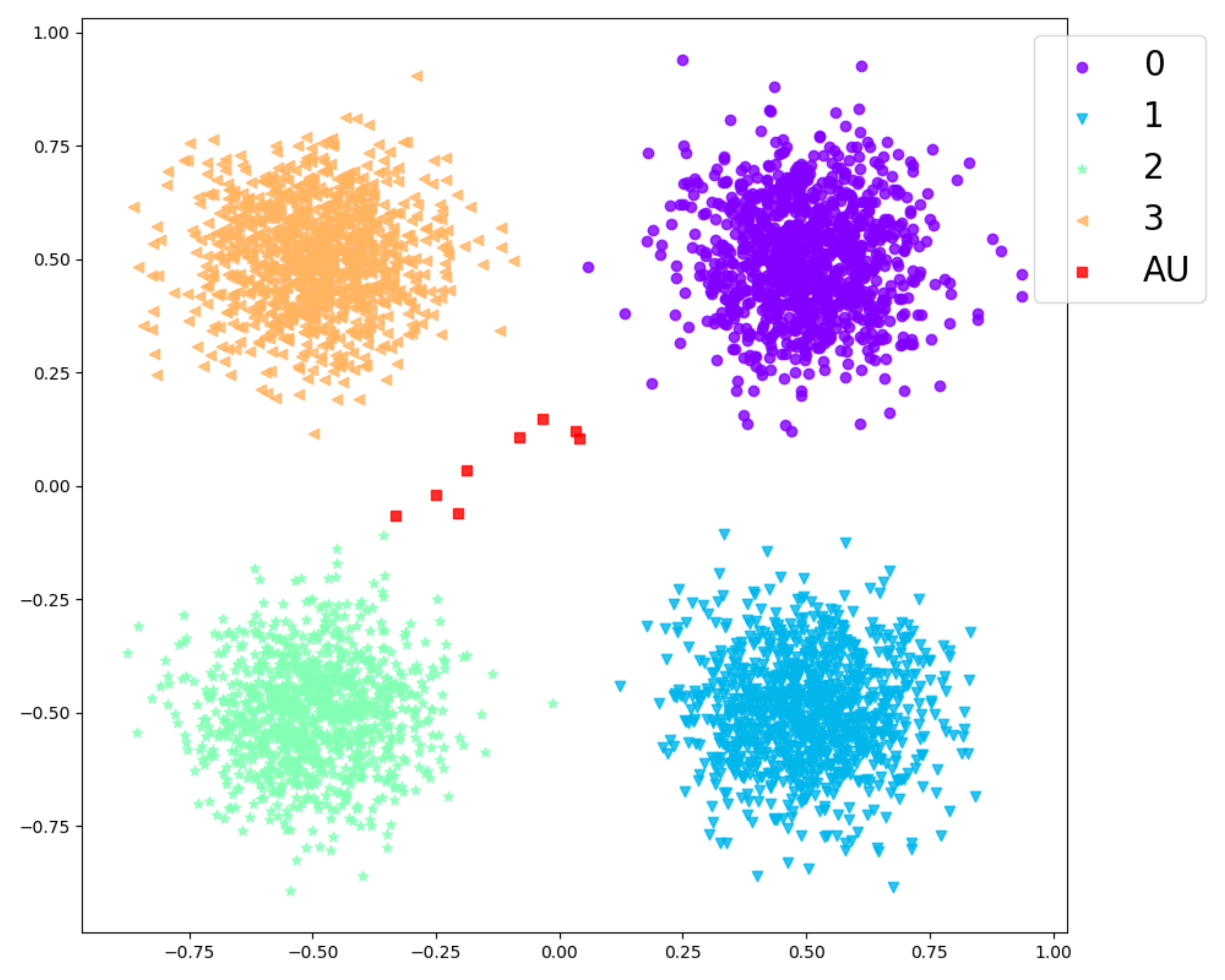}}
\subfloat[Epoch 23]{\includegraphics[width=0.3\linewidth]{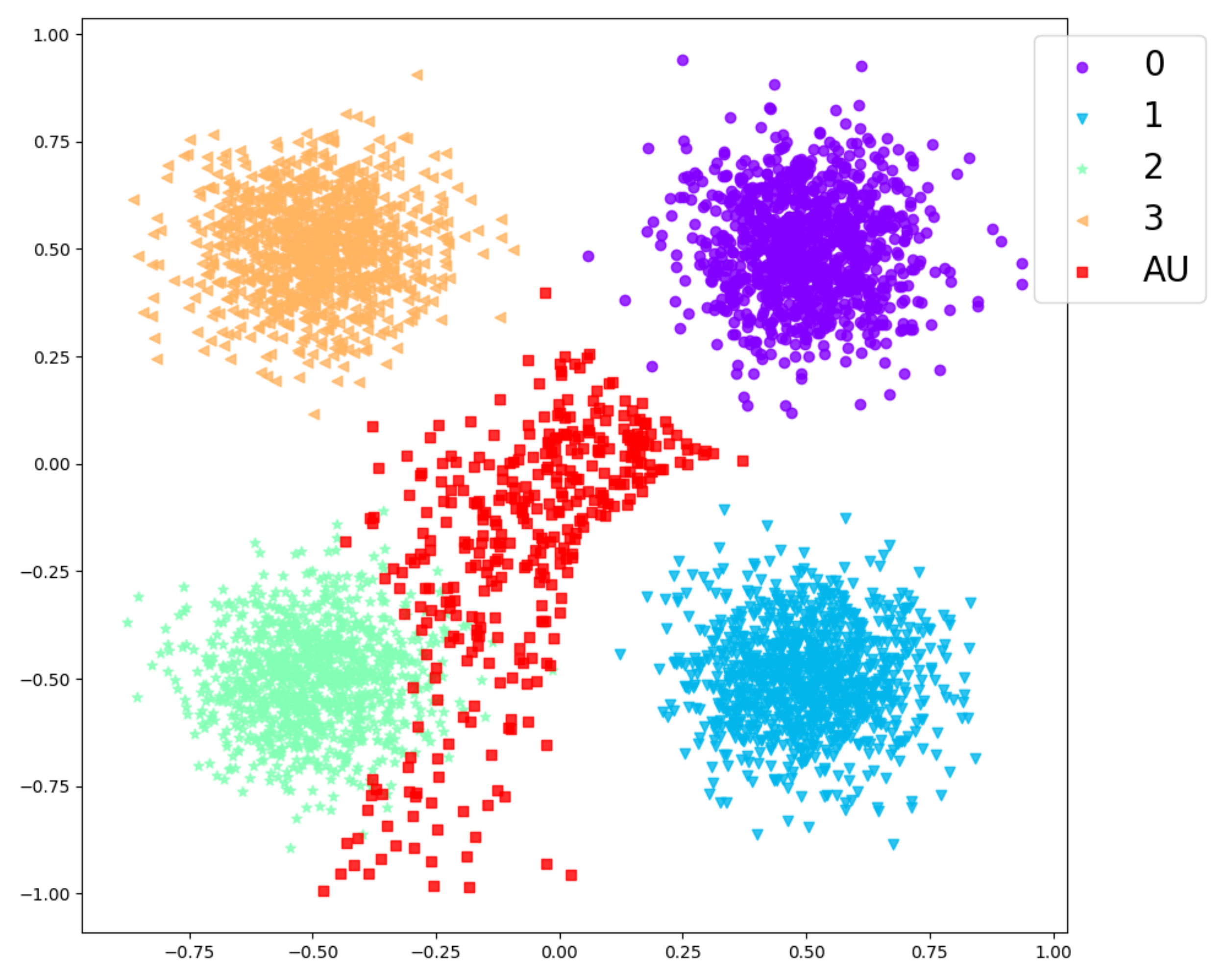}}
\subfloat[Epoch 27]{\includegraphics[width=0.3\linewidth]{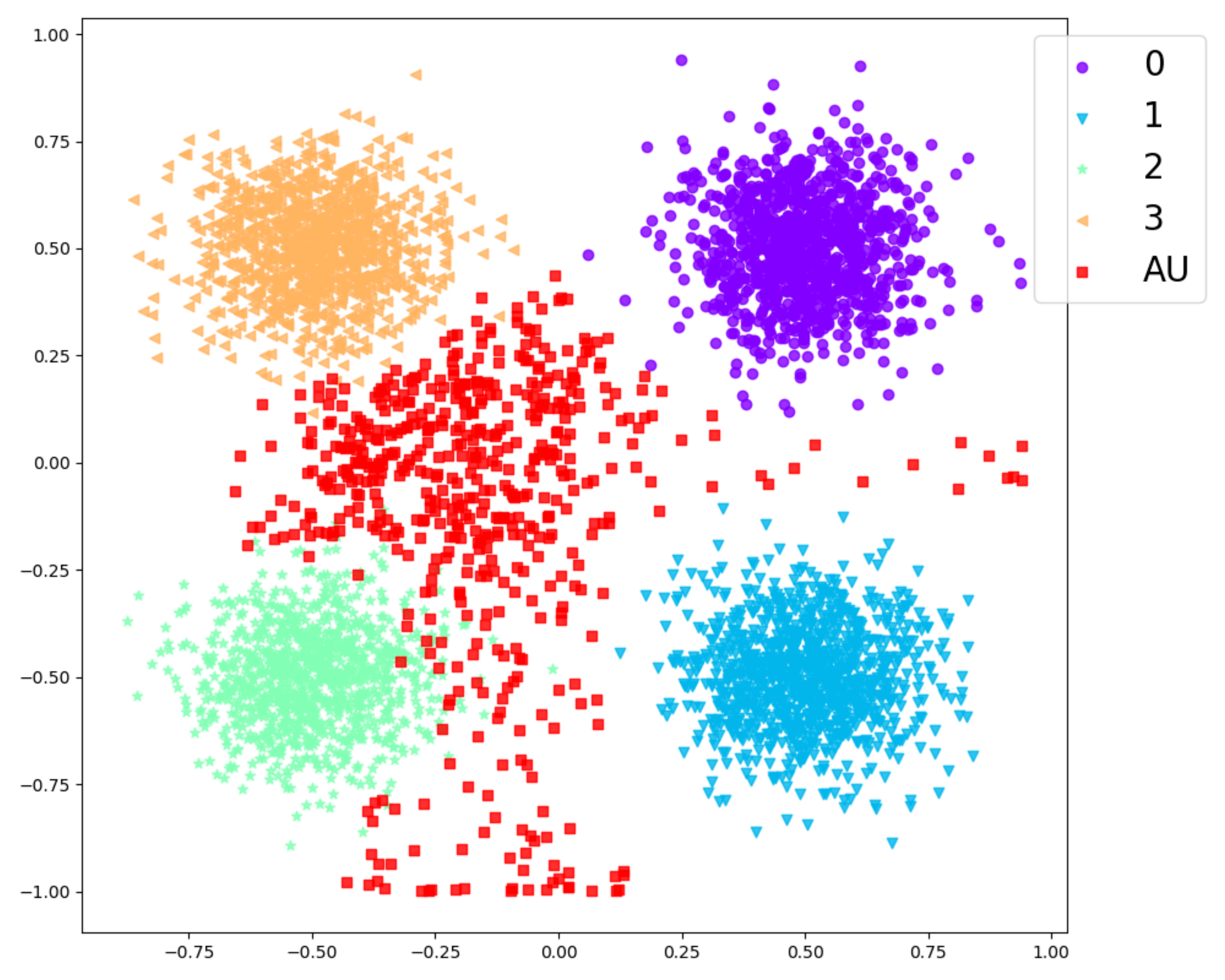}}
\par
\subfloat[Epoch 37]{\includegraphics[width=0.3\linewidth]{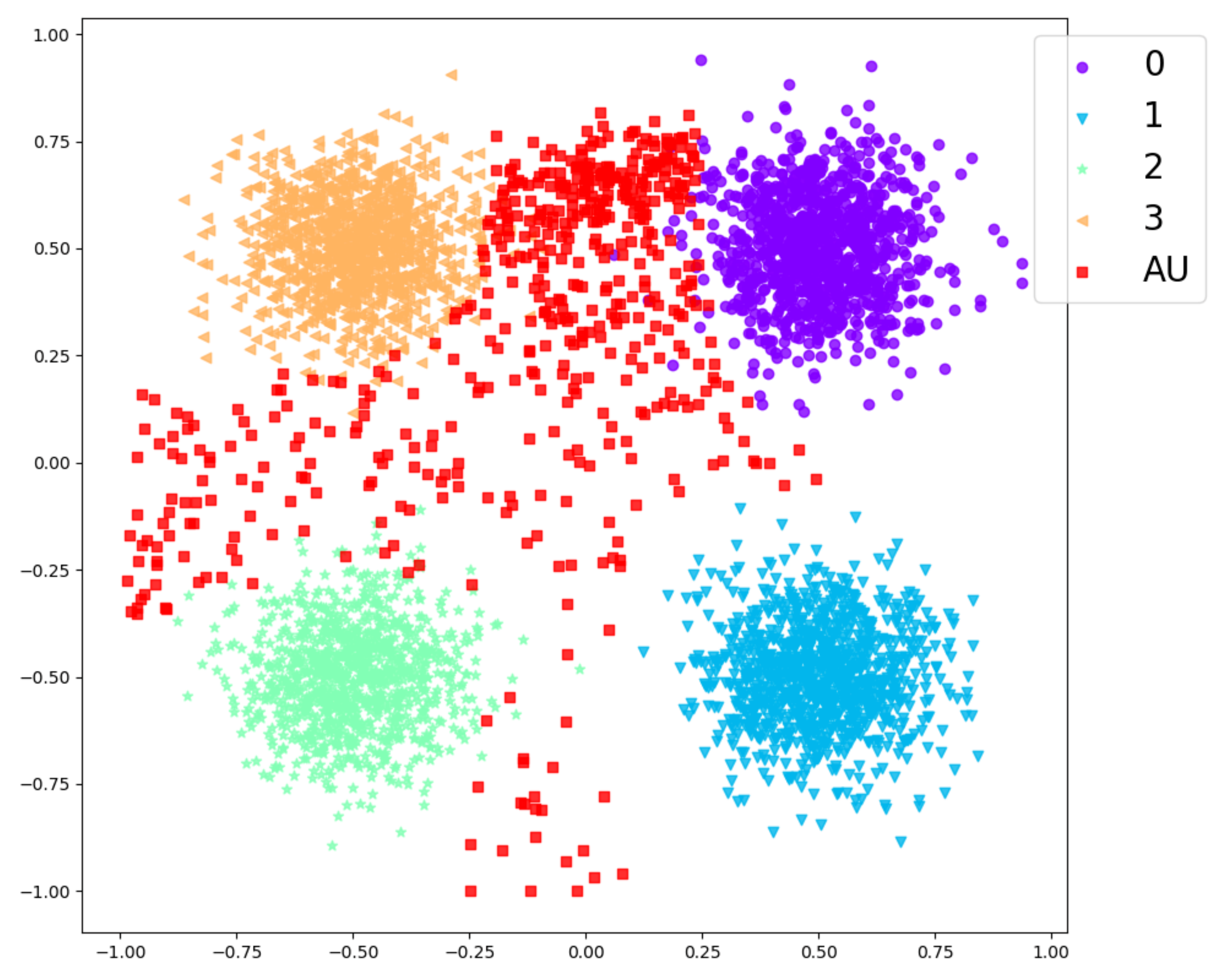}}
\subfloat[Epoch 47]{\includegraphics[width=0.3\linewidth]{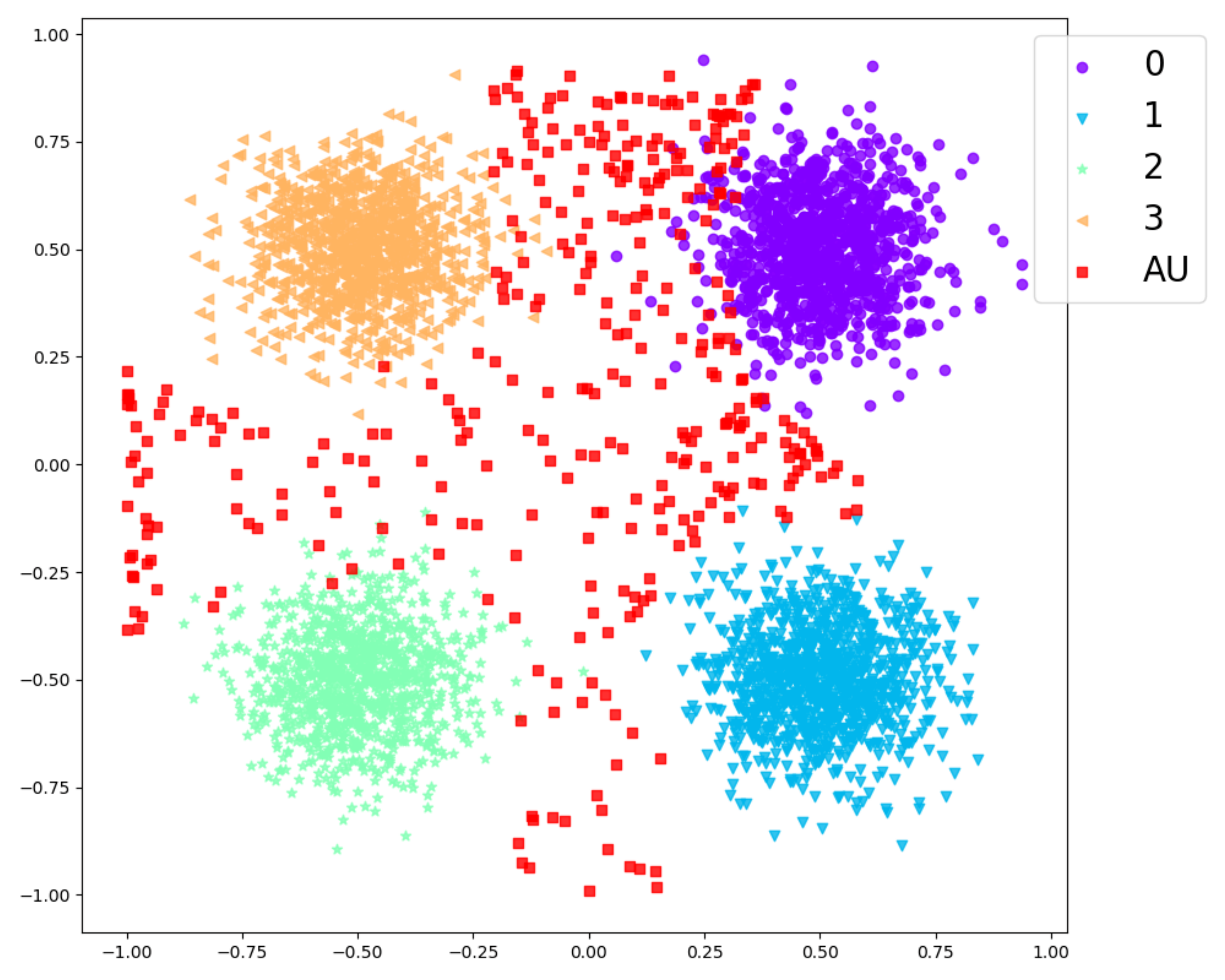}}
\subfloat[Epoch 57]{\includegraphics[width=0.3\linewidth]{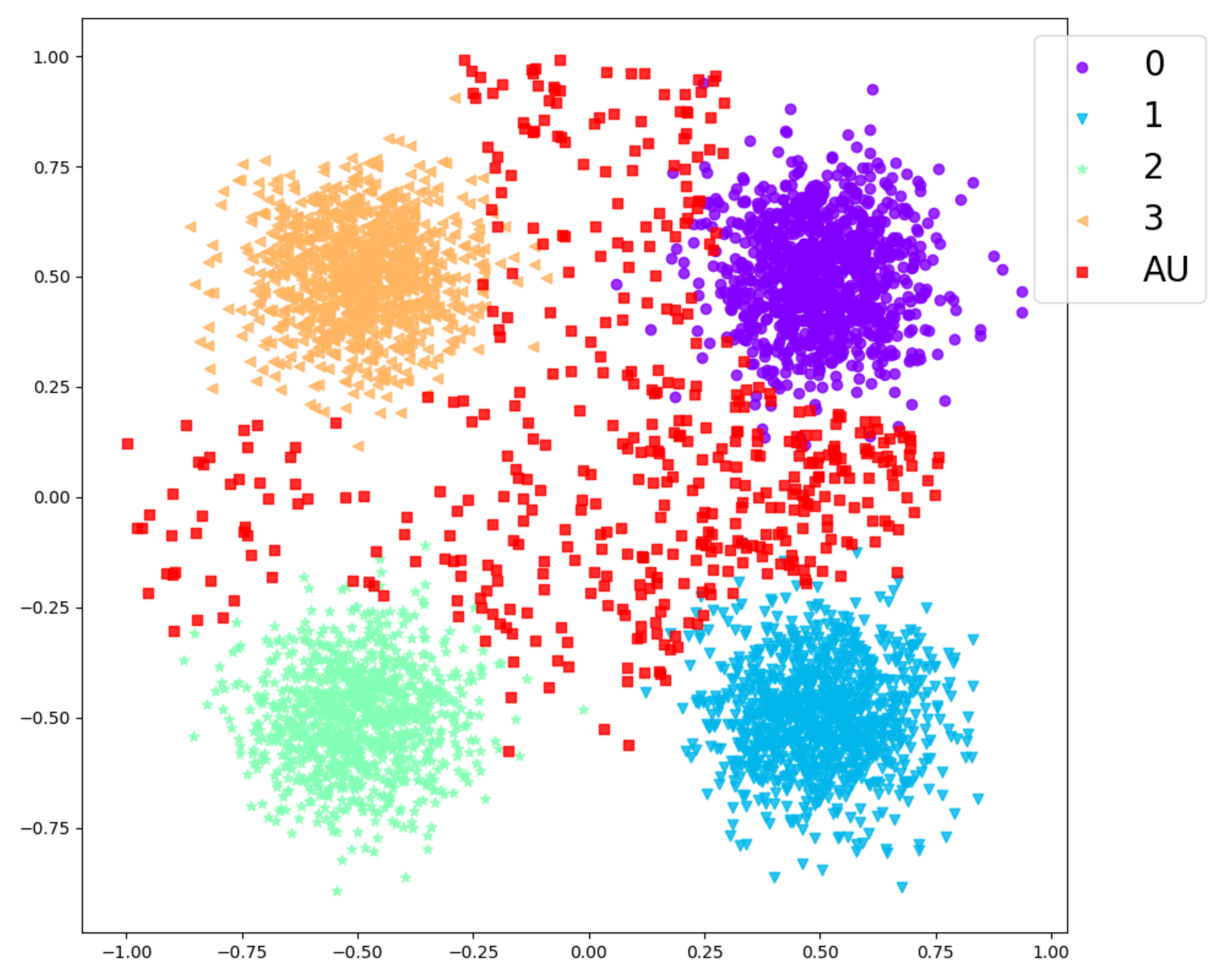}}
\caption{Generated active unknown samples after each epoch. AU denotes active unknown.}
\label{fig_8}
\end{minipage}
\end{figure*}

The T/E/S networks were trained on the MNIST dataset, and Fig. \ref{fig_5} shows the change in examples generated by the explorer. The examples were produced by feeding 15 fixed latent vectors after every training epoch. Interestingly, the generator produced digit-like images consisting of an unknown character and a black background after a few epochs. In addition, the generator continuously changed the pattern of the synthetic samples throughout the training period. Thus, in T/E/S learning, the explorer can provide the student with a variety of unknown-like examples that change slightly with each iteration. It was also confirmed that T/E/S learning converged even with learning open set examples that change continuously. First, Fig. \ref{fig_6} shows that the explorer easily reaches a learning equilibrium state, where the generator and the discriminator compete with almost equal strength. Similarly, Fig. \ref{fig_7} shows that the loss of the student converges well while training both real and fake samples and that the generator easily makes unknown-like samples to satisfy the student.

The toy example dataset consisting of four classes, each of which contains 1,000 samples, was generated. We applied T/E/S learning to this toy dataset. After every epoch of student and explorer learning, 1,000 examples were generated by the generator. Among the generated samples, active unknown samples were selected and plotted, as shown in Fig. \ref{fig_8}. In the early stages of training, the generator produced only fake samples with very high confidence for the known classes. After 19 epochs, active unknown samples were generated for the first time. At that time, only a few examples existed in the central region. As network training continued, more examples were produced around the place where active unknown samples were in the early epochs. The active unknown samples then continue to be repositioned. In explorer learning, the discriminator is trained to determine the generated samples as fake, and the generator is trained to deceive the trained discriminator alternately. Thus, the generator is forced to continuously change the distribution of the generated examples. This change in the distribution of generated samples helps the student network learn a variety of open set examples.

There is the risk that active unknown samples can violate the areas of the known classes. However, it is unlikely that the student's OSR performance decreases because the generated samples cannot stay in the same overlapping region. After the overlapped active unknown samples move away, the student relearns the known samples in the overlapped region. Rather, active unknown samples close to known class areas help the student network reduce open space by building stricter decision boundaries.

\begin{figure}[h]\centering
  \includegraphics[width=0.65\linewidth]{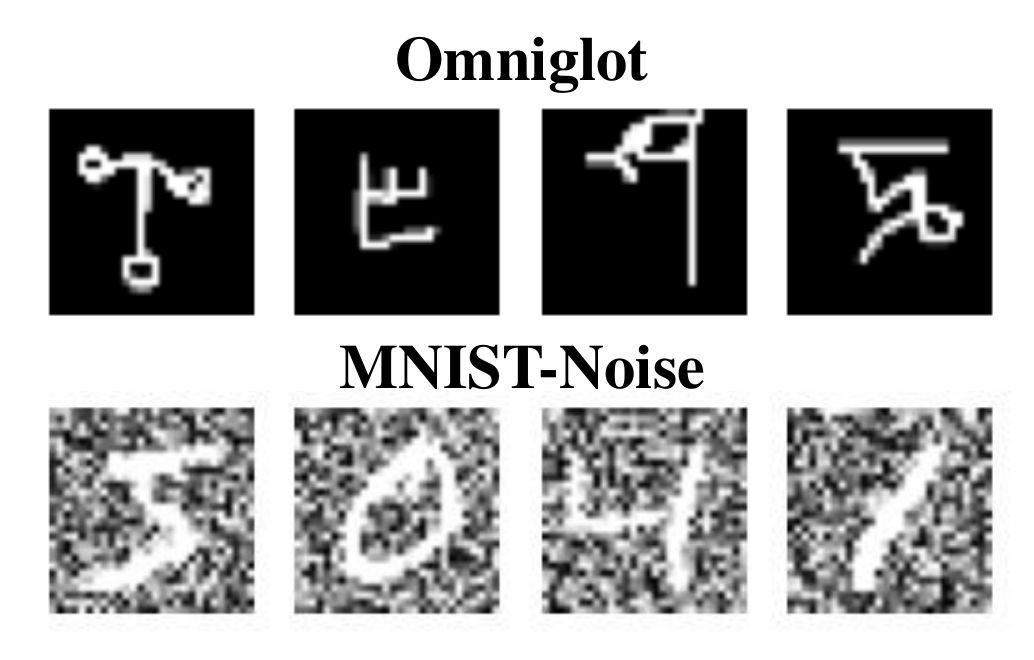}
  \caption{Sample images from Omniglot and MNIST-Noise.}
  \label{fig_9}
\end{figure}

An experiment was designed to analyze whether the generated examples can represent the unknowns in the learned latent space of the student network. For this experiment, T/E/S learning was applied to the MNIST dataset. As an unknown class, we used two datasets of grayscale images, Omniglot \cite{Lake2015} and MNIST-Noise (see Fig. \ref{fig_9}). Here, MNIST-Noise is the dataset made by superimposing the test samples of the MNIST on a noise image set that are synthesized by independently sampling each pixel value from a uniform distribution on [0, 1]. We randomly selected 1,000 samples from each known class and each unknown dataset. In addition, 1,000 samples were generated by the explorer.

\vspace{-10pt}

\begin{figure}[h]
\centering
\subfloat[]{\includegraphics[width=\linewidth]{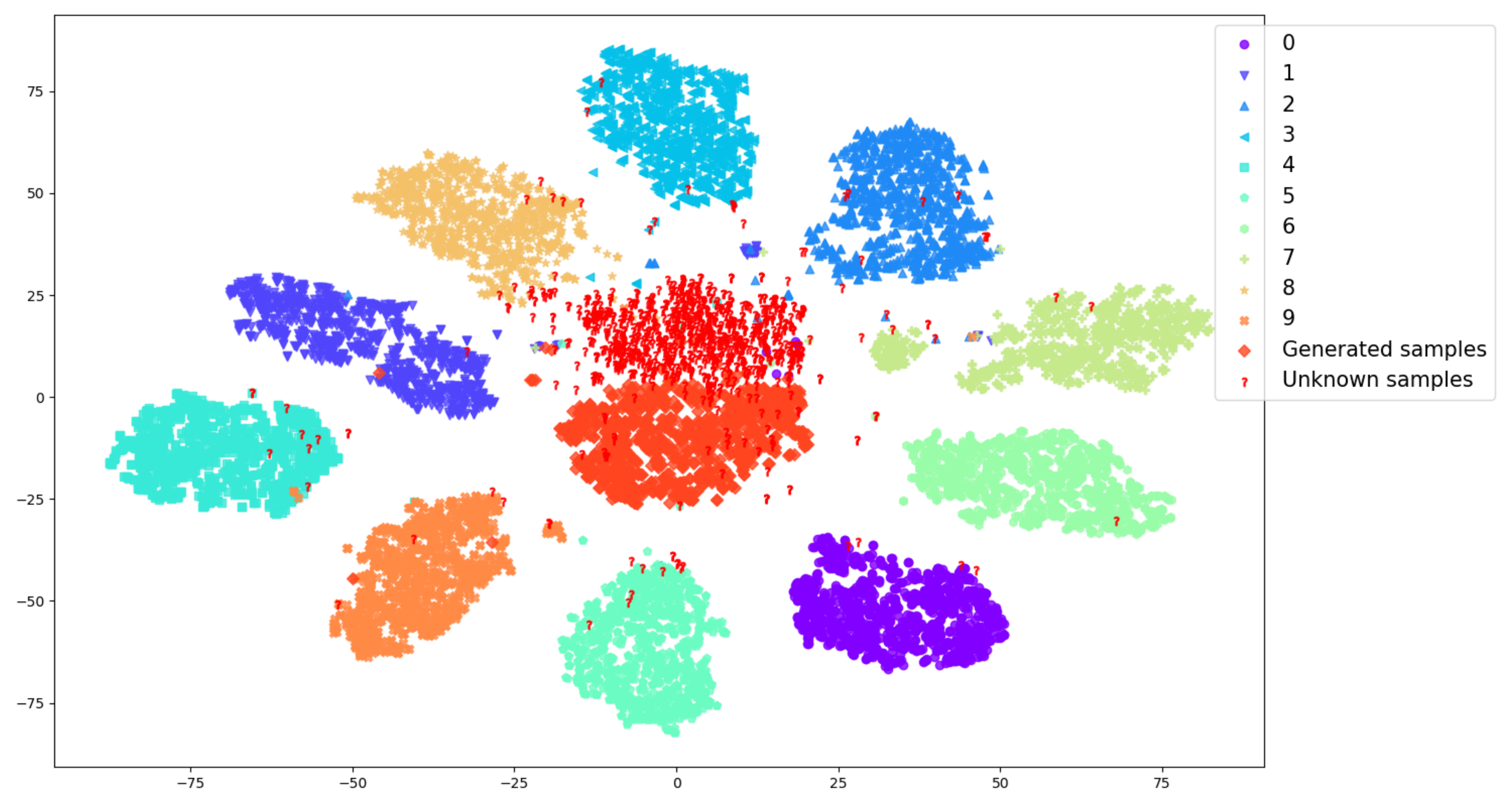}}
\par
\subfloat[]{\includegraphics[width=\linewidth]{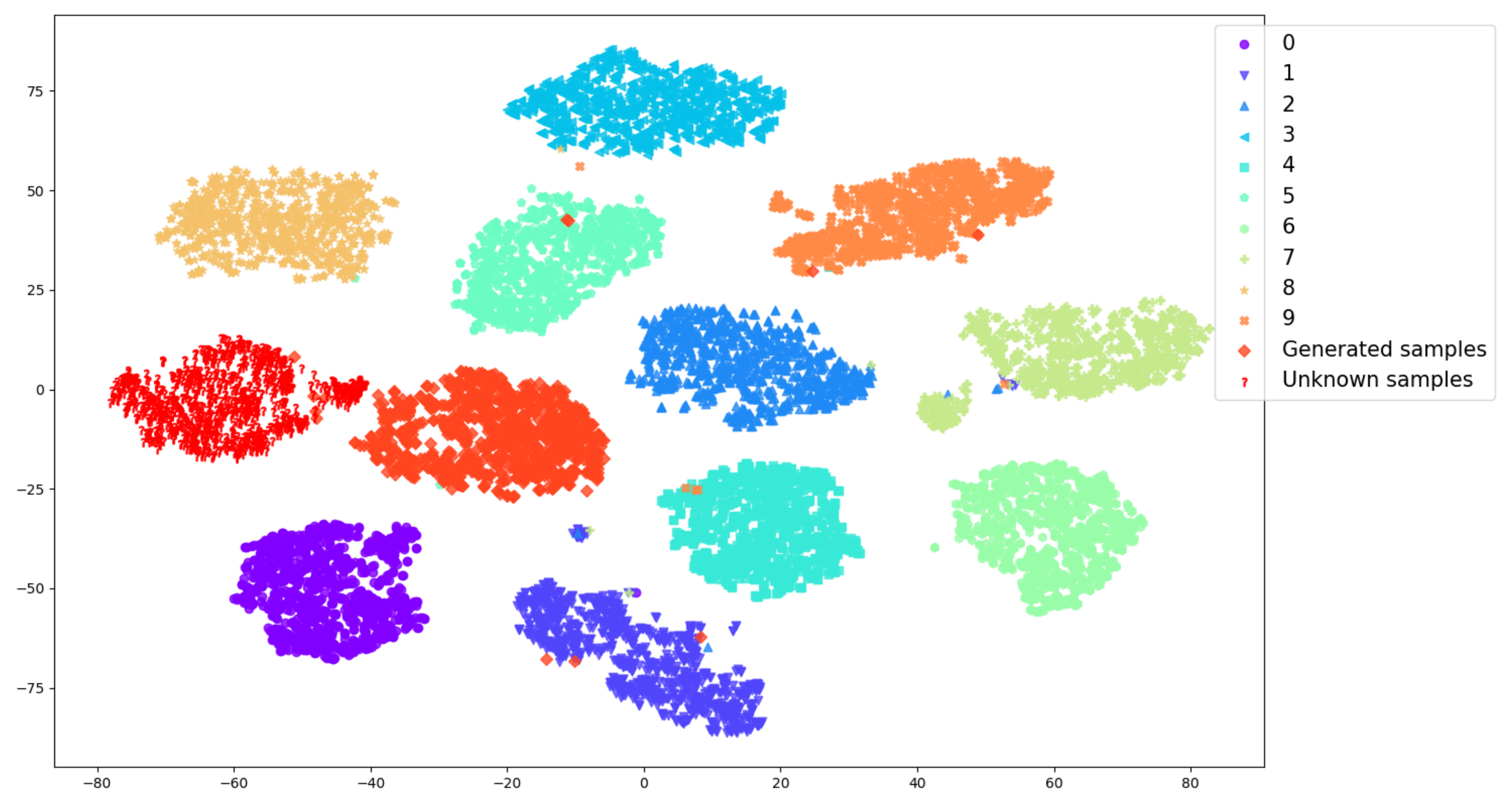}}
\caption{t-SNE visualization of known, generated, and unknown samples. (a) Omniglot and (b) MNIST-Noise were used as unknown samples.}
\label{fig_10}
\vspace{-10pt}
\end{figure}

The learned latent representations of the known samples, the unknown samples, and the generated samples were visualized with t-distributed stochastic neighbor embedding (t-SNE) \cite{Maaten2008}, as shown in Fig. \ref{fig_10}. The generated samples are clearly separated from the known classes, showing that the generator of the explorer mostly produces samples located in open space. Specifically, for Omniglot, most of the unknown samples are very close to the generated samples, creating overlapping regions. On the other hand, only a few samples are close to known class samples. In addition, the MNIST-Noise samples are closest to the cluster of generated samples, even if they look like MNIST samples. This is because T/E/S learning builds very tight class-specific decision boundaries to discriminate similar looking fake samples. The results show that the explorer can generate unknown-like samples.

\subsection{Ablation Study}
\begin{figure*}[t]
\centering
\subfloat[]{\includegraphics[width=0.33\linewidth]{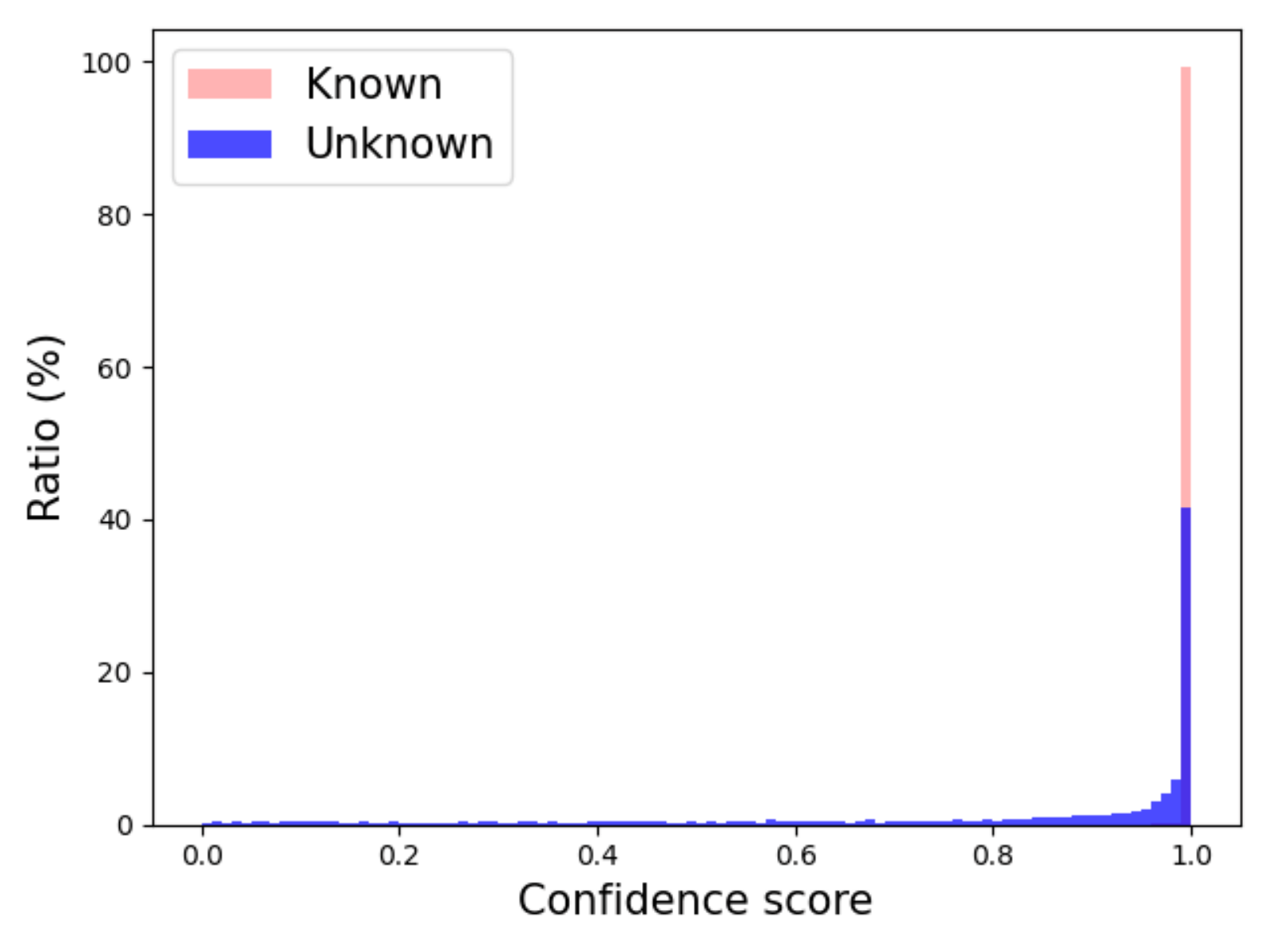}}
\subfloat[]{\includegraphics[width=0.33\linewidth]{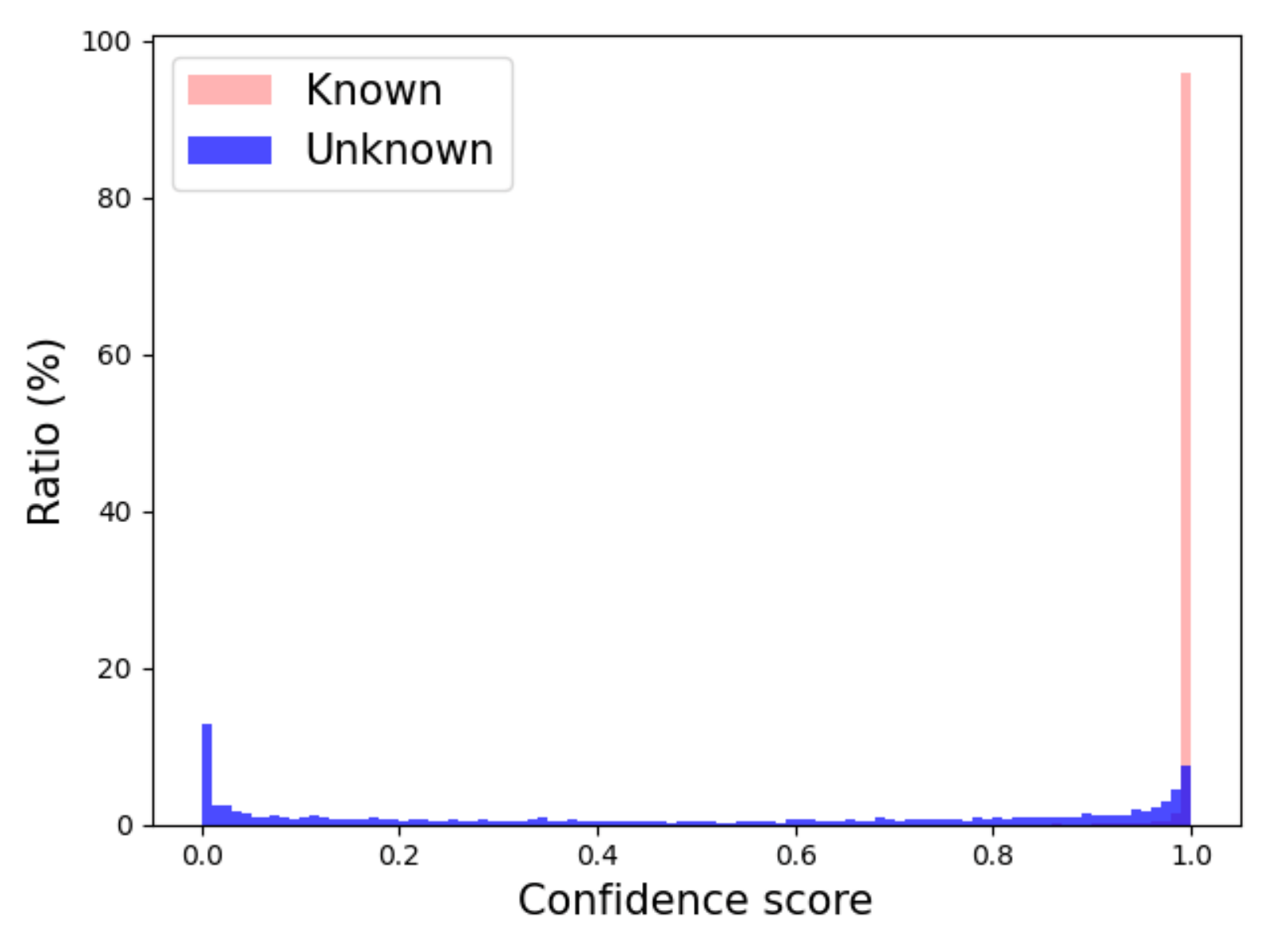}}
\subfloat[]{\includegraphics[width=0.33\linewidth]{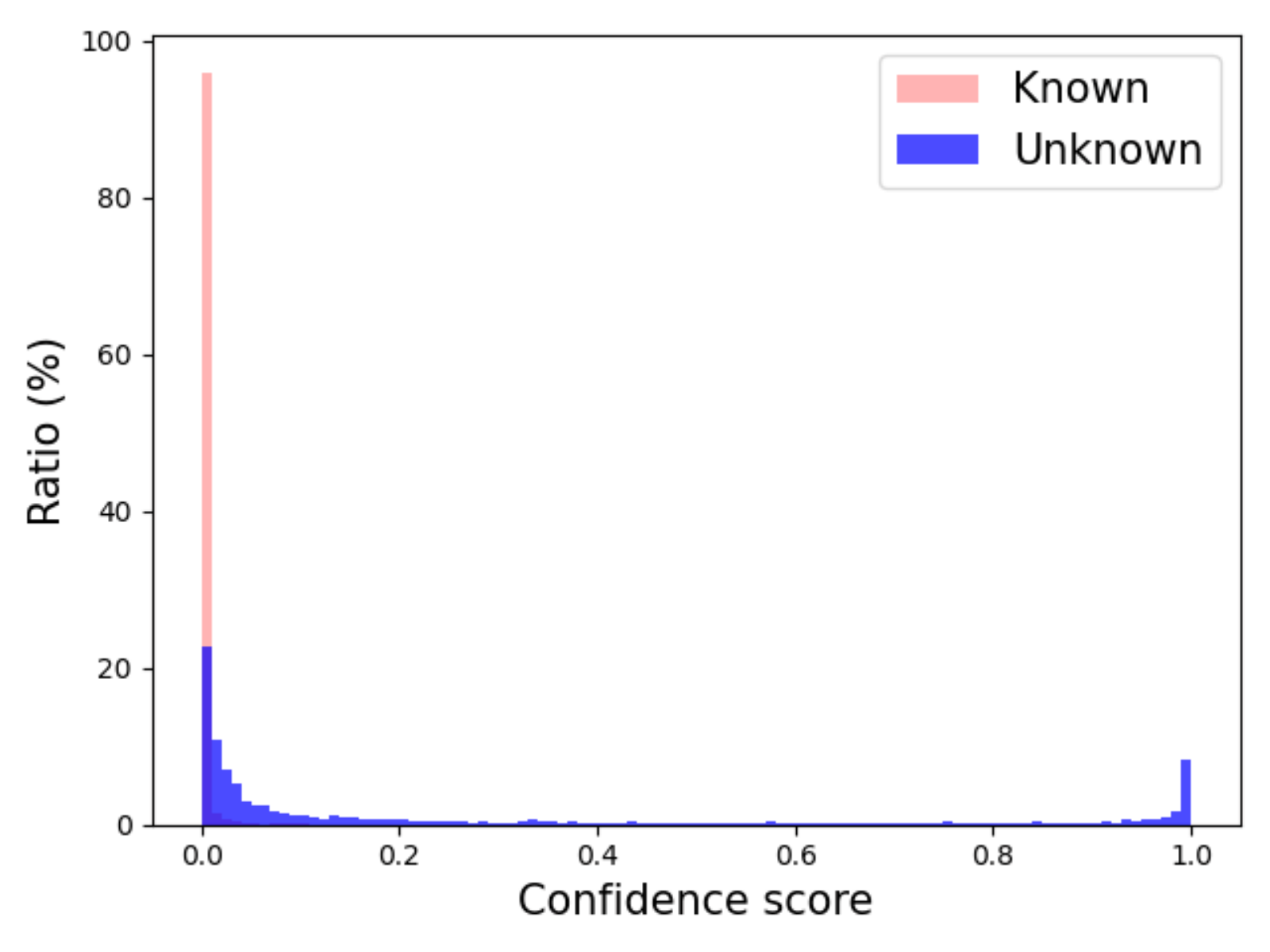}}
\caption{Confidence score distributions on MNIST. For (a) the CNN-OVRN and (b) the student network, the maximum sigmoid score among the known classes was used as the confidence score. (c) shows the distribution of uncertainty extracted by the student.}
\label{fig_11}
\vspace{-10pt}
\end{figure*}

We first conducted a qualitative analysis. The MNIST dataset was partitioned into six known classes $(0\sim 5)$ and four unknown classes $(6\sim 9)$. We trained a CNN with OVRNs (CNN-OVRN), which only applied the structure of the student network, and T/E/S networks on the known classes’ training dataset. The difference between Fig. \ref{fig_11}(a) and (b) shows that T/E/S learning can reduce overgeneralization significantly by providing low confidence scores to unknowns. In addition, Fig. \ref{fig_11}(c) shows that most unknown samples produced significantly higher uncertainty scores than known samples, even though real unknown samples were never trained to have high uncertainty. Specifically, approximately 14.2\% of unknown samples scored higher than 0.9. This reveals that T/E/S learning can infer some information about the unknown without direct training.

\begin{figure*}[b]
\centering
\subfloat[]{\includegraphics[width=0.33\linewidth]{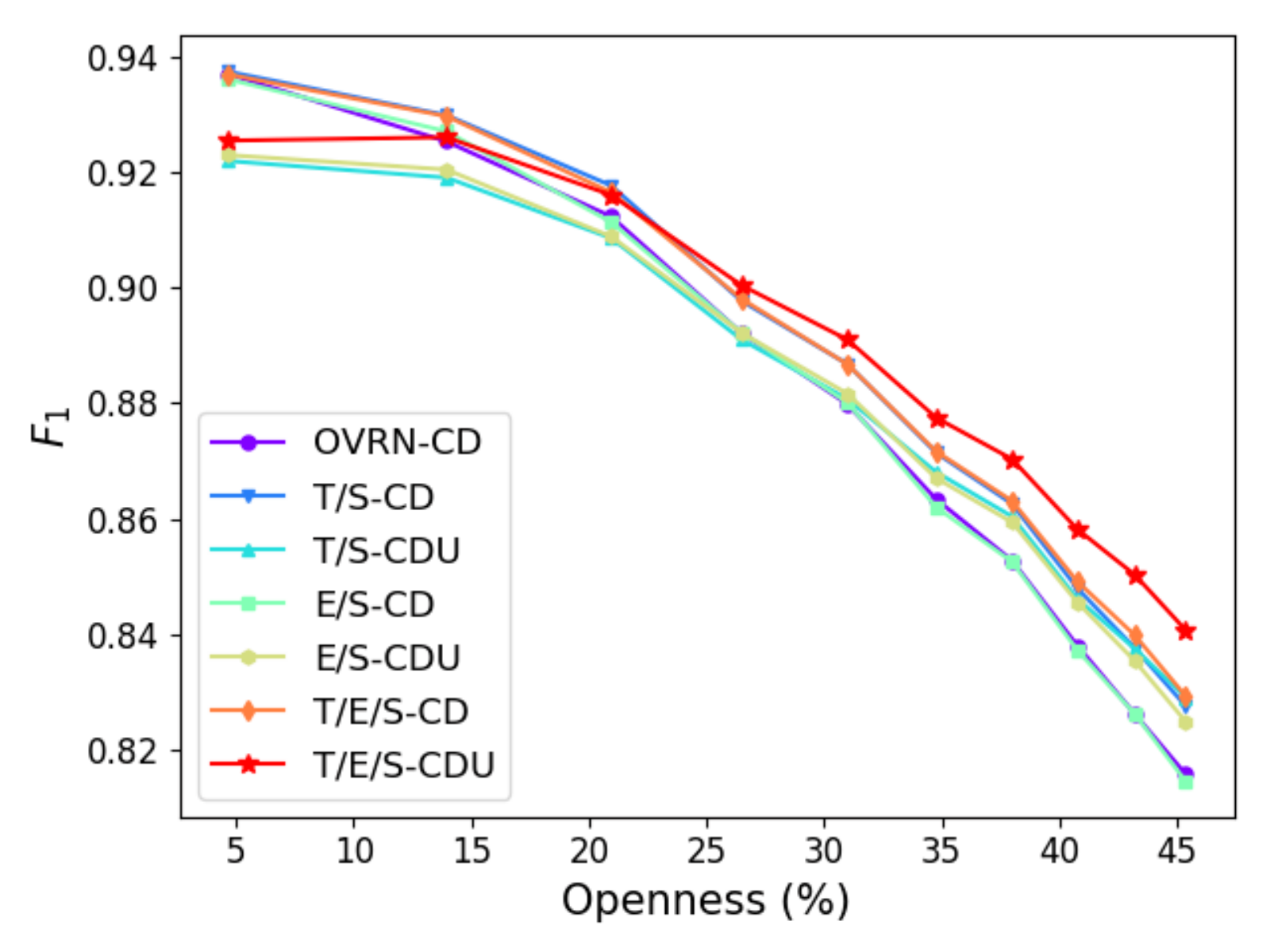}}
\subfloat[]{\includegraphics[width=0.33\linewidth]{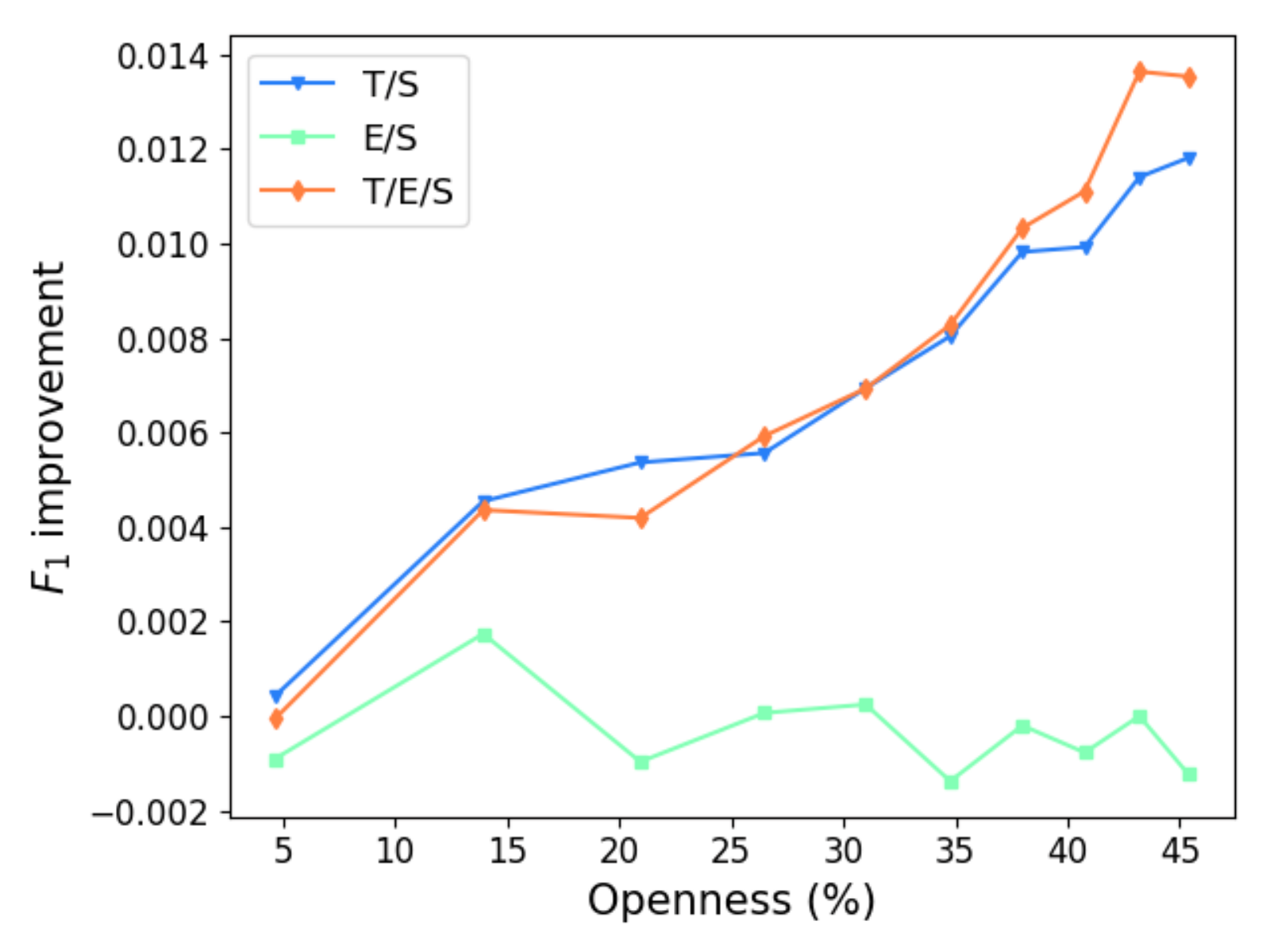}}
\subfloat[]{\includegraphics[width=0.33\linewidth]{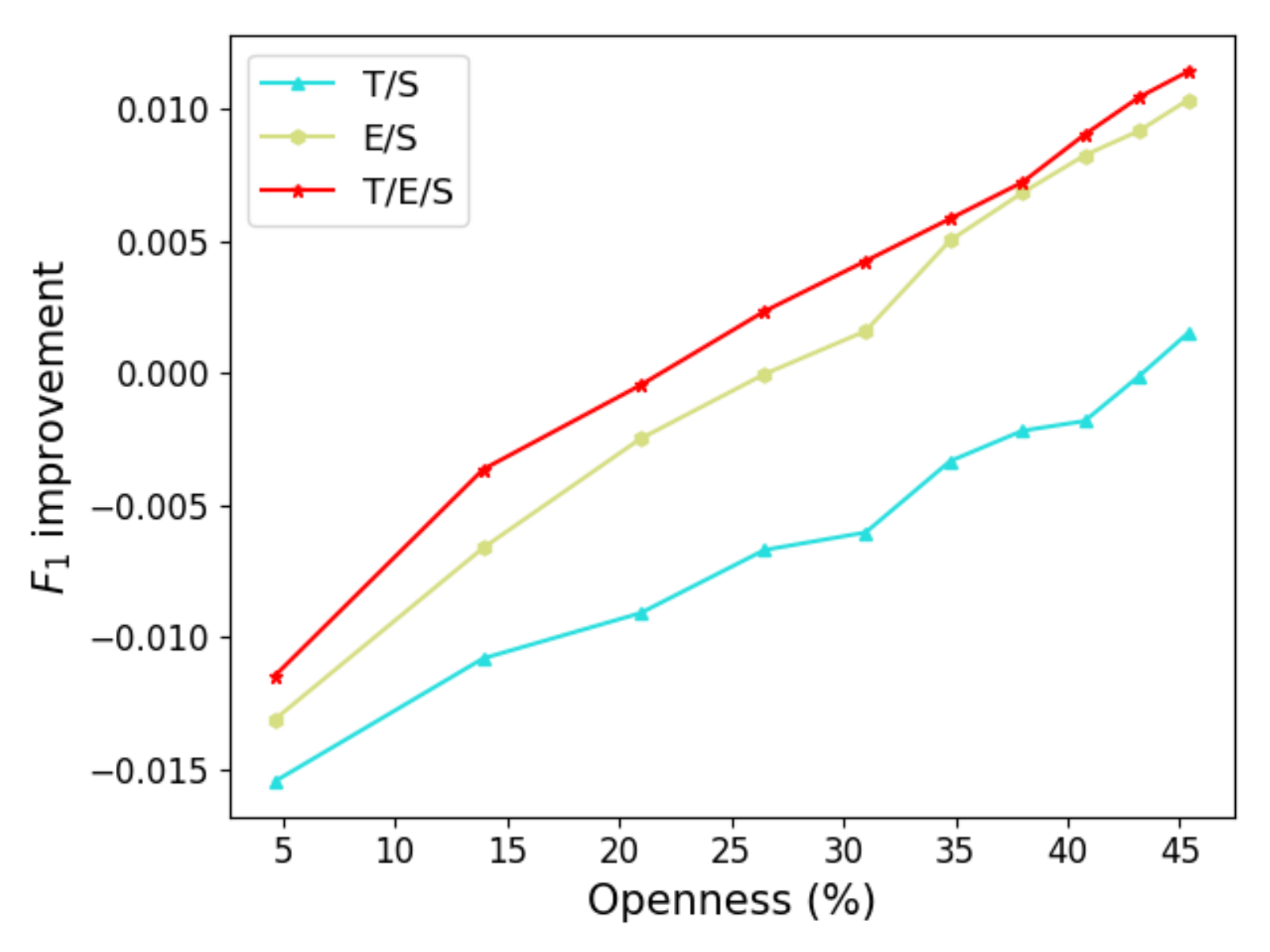}}
\caption{F1 score with seven baselines (a) and each component’s contribution (b)/(c) when MNIST and EMNIST were used as known classes and unknown classes, respectively. In (b), T/S, E/S, and T/E/S each represent the performance improvements obtained by introducing the teacher, explorer, and both networks in addition to OVRN-CD. (c) shows the performance improvements obtained by introducing uncertainty thresholds to T/S, E/S, and T/E/S.}
\label{fig_12}
\vspace{-10pt}
\end{figure*}

As a quantitative analysis, the effects of the network compositions were analyzed. All the baseline models in this section used the collective decision scores for OSR. Specifically, the following seven baselines were compared:
\begin{enumerate}[1)]
\item OVRN-CD: This baseline is the CNN-OVRN with the collective decision method.
\item T/S-CD: A pretrained teacher network is additionally used to provide a student network with hints for the unknown samples by applying HE-KD.
\item T/S-CDU: In this baseline, the network composition and the training procedure are the same as in baseline 2, while the collective decision thresholds and the uncertainty threshold are used jointly for the decision rule.
\item E/S-CD: This baseline is different from baseline 2; an explorer is used to support a student by generating synthetic open set examples.
\item E/S-CDU: The uncertainty threshold is used in addition to baseline 4.
\item T/E/S-CD (proposed method): This baseline applies the proposed T/E/S learning method but does not take into account the uncertainty threshold.
\item T/E/S-CDU (proposed method): This baseline additionally introduces the uncertainty threshold into baseline 6.
\end{enumerate}
The performance was evaluated by the macroaverage F1-score ($F_1$) for known classes and “unknown.”

We adopted the two experimental settings suggested in \cite{Jang2020} for the quantitative ablation study. In the first experimental setting, we used 10 digit classes from the MNIST dataset as the known classes and 47 letter classes from EMNIST \cite{Cohen2017} as the unknown classes. In the second setting, we used four nonanimal classes from the CIFAR-10 dataset as known classes and 100 natural classes from the CIFAR-100 dataset as unknown classes. The OSR performance is significantly affected by the ratio of the unknown classes to the known classes. Thus, we set various openness values for each experimental setting. Here, openness is the measurement of how open the problem setting is and is defined as follows \cite{Scheirer2013}:
\begin{gather} 
\text{openness}=1-\sqrt{\frac{2C_T}{C_E+C_R}}, \label{eq12}
\end{gather}
where $C_T$ is the number of classes used in training, $C_E$ is the number of classes used in evaluation, and $C_R$ is the number of classes to be recognized. Specifically, we varied openness from 4.7\% to 45.4\% for the first setting and from 21.6\% to 72.8\% for the second setting.

Fig. \ref{fig_12}(a) shows the comparison results when MNIST and EMNIST were used. While T/E/S-CD provides the best performance in low openness settings, introducing the uncertainty threshold improves the robustness of recognition performance; hence, T/E/S-CDU has the best performance as the openness value increases. To analyze the contribution of each component, we designed two additional comparisons. First, we compared T/S-CD, E/S-CD, and T/E/S-CD with OVRN-CD to identify the contributions of introducing the teacher, the explorer, and both networks, as shown in Fig. \ref{fig_12}(b). In the figure, the introduction of the explorer did not improve the performance, but the introduction of the teacher increasingly improved the performance as openness increased. The result reveals that HE-KD alone can contribute to performance improvement.

\begin{figure*}[t]
\centering
\subfloat[]{\includegraphics[width=0.33\linewidth]{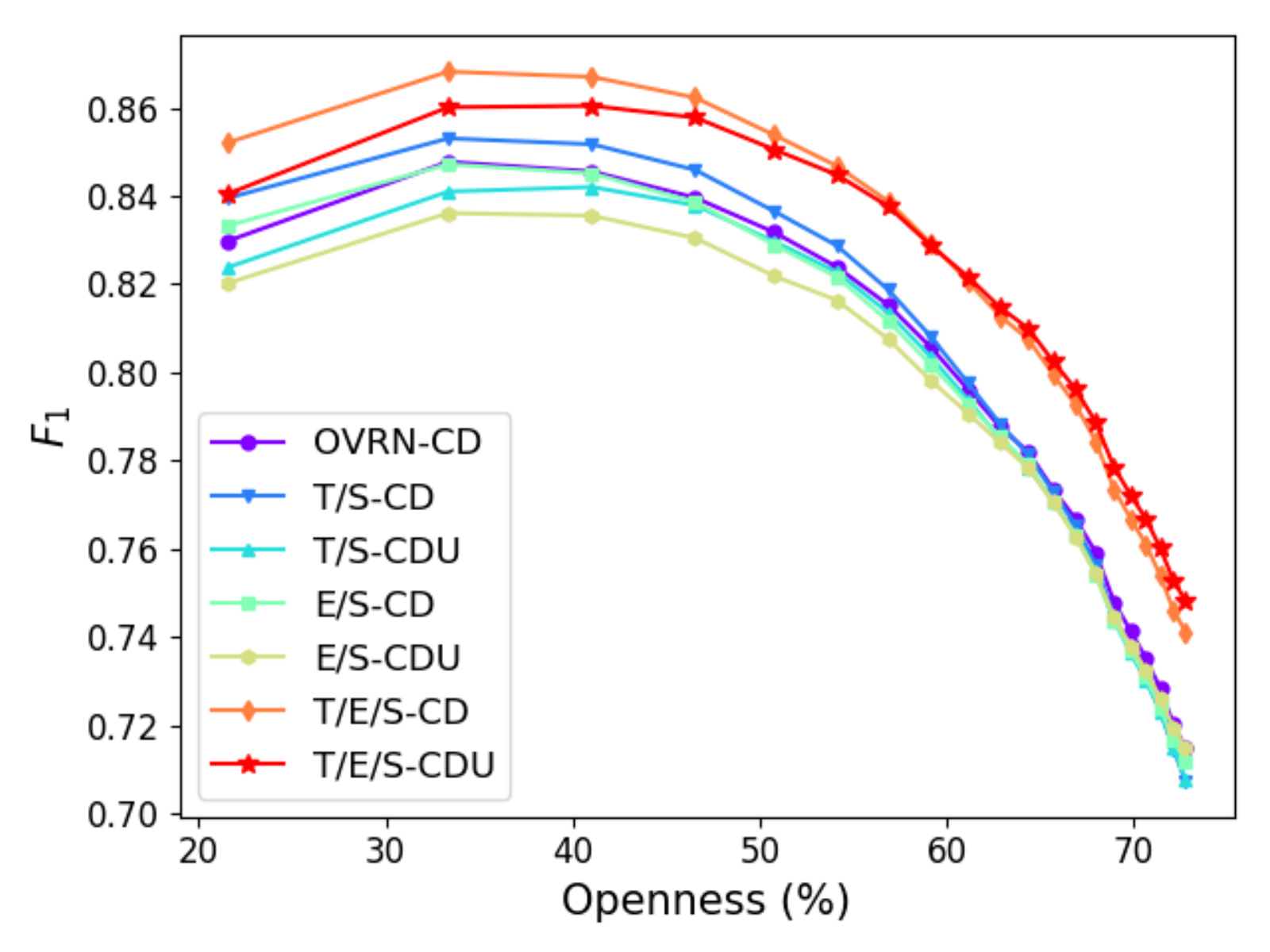}}
\subfloat[]{\includegraphics[width=0.33\linewidth]{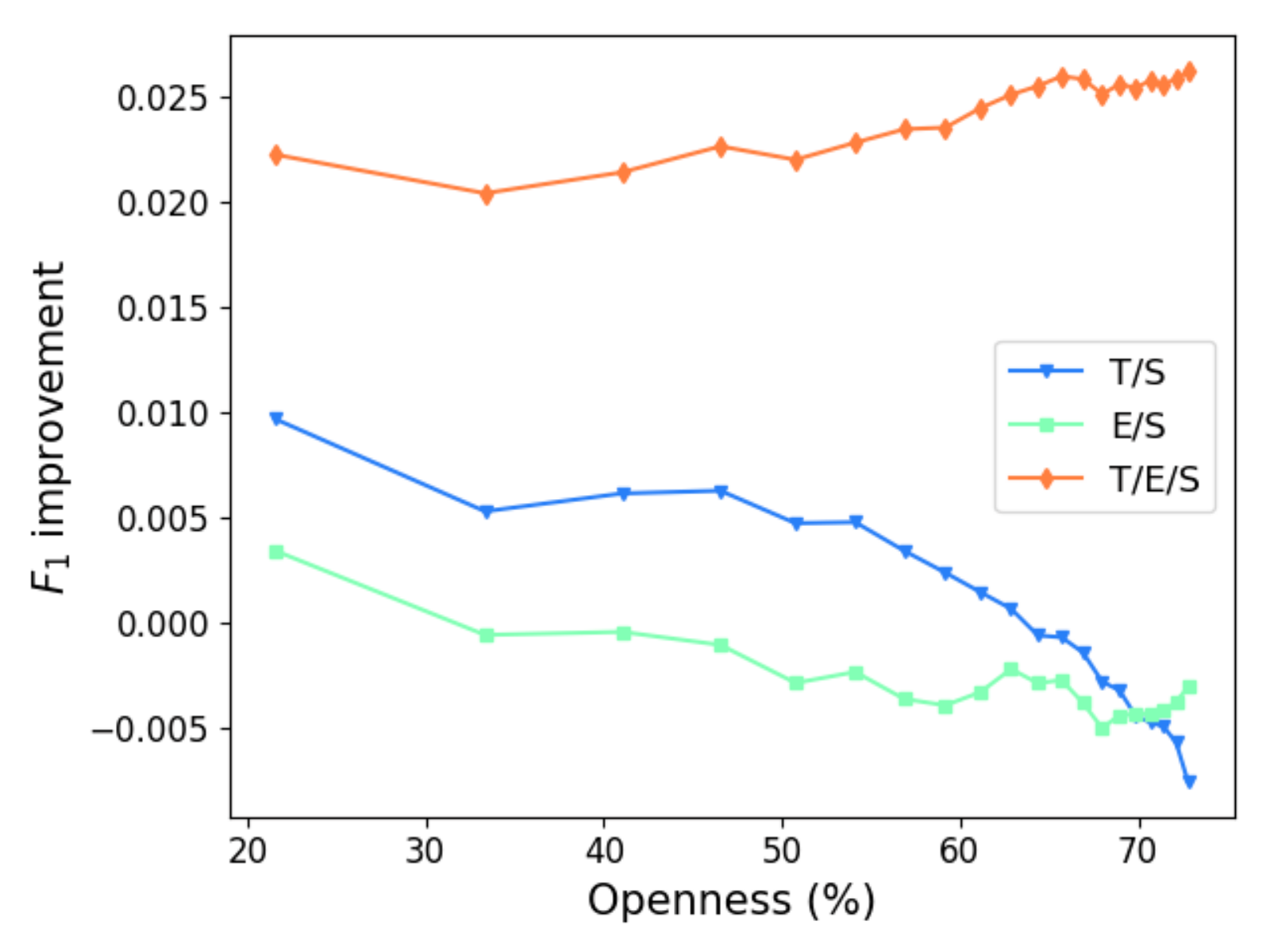}}
\subfloat[]{\includegraphics[width=0.33\linewidth]{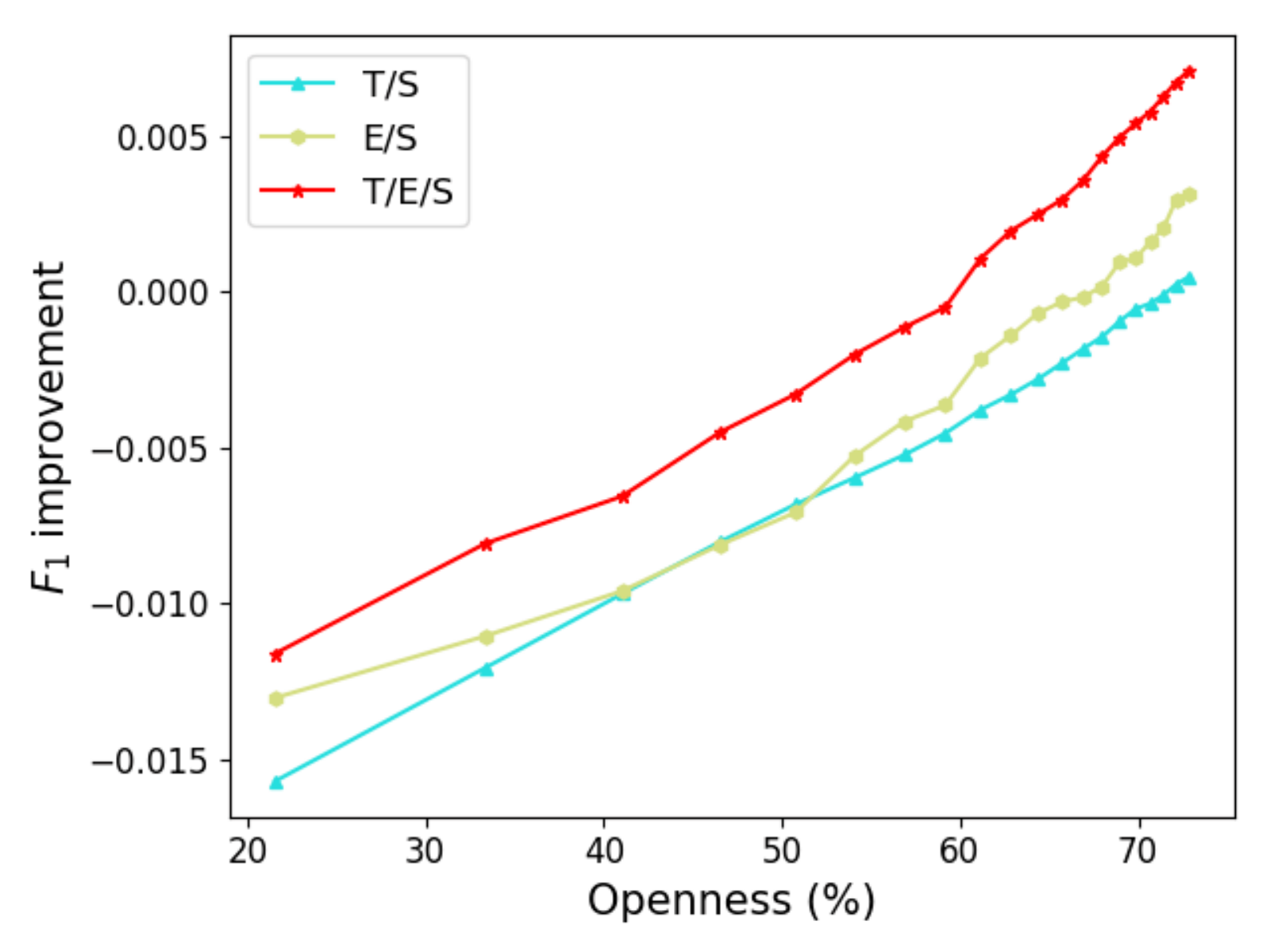}}
\caption{F1 score with seven baselines (a) and each component’s contribution (b)/(c) when CIFAR-10 and CIFAR-100 were used as known classes and unknown classes, respectively. (b) and (c) show the performance improvements obtained by introducing additional networks to OVRN-CD and introducing the uncertainty threshold, respectively.}
\label{fig_13}
\vspace{-10pt}
\end{figure*}

\begin{table*}[b]
\centering
\begin{threeparttable}
\caption{Unknown Detection Results}
\label{table_2}
\begin{tabular}{cc|cccccc|c}
\hline								
Approach & Method &MNIST & SVHN & CIFAR-10 & CIFAR+10 & CIFAR+50 & Tiny-ImageNet & Avg. \\
\hline	
 & SoftMax & 0.978 & 0.886 & 0.677 & 0.816 & 0.805 & 0.577 & 0.790\\
 & OpenMax \cite{Bendale2016} & 0.981 & 0.894 & 0.695 & 0.817 & 0.796 & 0.576 & 0.793\\
 & G-OpenMax \cite{Ge2017} & 0.984 & 0.896 & 0.675 & 0.827 & 0.819 & 0.580 & 0.797\\
One-stage & OSRCI \cite{Neal2018} & 0.988 & 0.910 & 0.699 & 0.838 & 0.827 & 0.586 & 0.808\\
 & DOC \cite{Shu2017} & 0.986 & 0.937 & 0.885 & 0.880 & 0.882 & 0.725 & 0.883\\
 & OVRN-CD \cite{Jang2020} & 0.989 & 0.941 & 0.903 & 0.907 & 0.902 & 0.730 & 0.895\\
 & T/E/S-CD(U) & \underline{0.993} & \underline{\textbf{0.953}} & \underline{\textbf{0.905}} & \underline{0.951} & \underline{0.924} & \underline{0.733} & \underline{0.910}\\
\hline
 & MLOSR \cite{Oza2019a} & 0.989 & 0.921 & 0.845 & 0.895 & 0.877 & 0.718 & 0.874\\
Two-stage & C2AE \cite{Oza2019} & 0.989 & 0.922 & 0.895 & 0.955 & 0.937 & 0.748 & 0.908\\
 & CGDL \cite{Sun2020} & \underline{\textbf{0.994}} & \underline{0.935} & \underline{0.903} & \underline{\textbf{0.959}} & \underline{\textbf{0.950}} & \underline{\textbf{0.762}} & \underline{\textbf{0.917}}\\
\hline
\end{tabular}
    \begin{tablenotes}
      \small
      \item \textit{For each experimental setting, the best performing method is highlighted in bold and the best method in each approach is underlined in the table.}
    \end{tablenotes}
\end{threeparttable}
\vspace{-10pt}
\end{table*}

\begin{table*}[t]
\caption{OSR Performance Comparison Results on CIFAR-10 with Various “Unknown” Datasets}
\label{table_3}
\centering
\begin{tabular}{l|c|c|c|c|c}
\hline\hline
Method & ImageNet-crop & ImageNet-resize & LSUN-crop & LSUN-resize & Avg. \\
\hline									
Softmax & 0.639 & 0.653 & 0.642 & 0.647 & 0.645\\
OpenMax \cite{Bendale2016} & 0.660 & 0.684 & 0.657 & 0.668 & 0.667\\
LadderNet+Softmax \cite{Yoshihashi2019} & 0.640 & 0.646 & 0.644 & 0.647 & 0.644\\
LadderNet+OpenMax \cite{Yoshihashi2019} & 0.653 & 0.670 & 0.652 & 0.659 & 0.659\\
DHRNet+Softmax \cite{Yoshihashi2019} & 0.645 & 0.649 & 0.650 & 0.649 & 0.648\\
DHRNet+OpenMax \cite{Yoshihashi2019} & 0.655 & 0.675 & 0.656 & 0.664 & 0.663\\
CROSR \cite{Yoshihashi2019} & 0.721 & 0.735 & 0.720 & 0.749 & 0.731\\
DOC \cite{Shu2017} & 0.760 & 0.753 & 0.748 & 0.764 & 0.756\\
OVRN-CD \cite{Jang2020} & 0.835 & 0.825 & 0.846 & \textbf{0.839} & 0.836\\
MLOSR \cite{Oza2019a} (Two-stage)  & 0.837 & 0.826 & 0.783 & 0.801 & 0.812\\
CGDL \cite{Sun2020} (Two-stage)  & 0.840 & \textbf{0.832} & 0.806 & 0.812 & 0.823\\
T/E/S-CD (Ours) & \textbf{0.852} & 0.816 & \textbf{0.851} & 0.837 & \textbf{0.839}\\
T/E/S-CDU (Ours) & 0.843 & 0.808 & 0.843 & 0.828 & 0.831\\
\hline
\end{tabular}
\vspace{-10pt}
\end{table*}

\begin{table}[h]
\caption{OSR Performance Comparison Results on MNIST with Various “Unknown” Datasets}
\label{table_4}
\centering
\begin{tabular}{c|c|c|c}
\hline\hline
Method & Omniglot & MNIST-Noise & Noise\\
\hline									
Softmax & 0.592 & 0.641 & 0.826\\
OpenMax \cite{Bendale2016} & 0.680 & 0.720 & 0.890\\
LadderNet+Softmax \cite{Yoshihashi2019} & 0.588 & 0.772 & 0.828\\
LadderNet+OpenMax \cite{Yoshihashi2019} & 0.764 & 0.821 & 0.826\\
DHRNet+Softmax \cite{Yoshihashi2019} & 0.595 & 0.801 & 0.829\\
DHRNet+OpenMax \cite{Yoshihashi2019} & 0.780 & 0.816 & 0.826\\
CROSR \cite{Yoshihashi2019} & 0.793 & 0.827 & 0.826\\
DOC \cite{Shu2017} & 0.863 & 0.892 & 0.921\\
OVRN-CD \cite{Jang2020} & 0.918 & 0.926 & 0.953\\
CGDL \cite{Sun2020} (Two-stage)  & 0.850 & 0.887 & 0.859\\
T/E/S-CD (Ours) & \textbf{0.940} & \textbf{0.972} & \textbf{0.972}\\
T/E/S-CDU (Ours)  & 0.938 & 0.964 & 0.964\\
\hline
\end{tabular}
\vspace{-10pt}
\end{table}

Second, we analyzed the addition of uncertainty threshold to each network composition by comparing T/S-CD, E/S-CD, and T/E/S-CD with T/S-CDU, E/S-CDU, and T/E/S-CDU, respectively, as shown in Fig. \ref{fig_12}(c). The figure shows that the uncertainty threshold contributes to the performance improvement in high openness only when the explorer is adopted. Even if the student in T/S learns the hints extracted by the teacher, this information alone cannot improve performance significantly. However, the hints can be used to guide the explorer to provide the student with more meaningful unknown-like examples, considering that the F1-score improvement in T/E/S is higher than that of E/S. In summary, the teacher and explorer networks have their own individual roles in improving OSR performance. In addition, introducing the two networks together created synergy.

Fig. \ref{fig_13}(a) shows that the proposed methods outperformed the other baselines when CIFAR-10 and CIFAR-100 were used. Specifically, T/E/S-CD achieved the best results for openness up to 59.2\%, while T/E/S-CDU achieved the best results for openness beyond 59.2\%. HE-KD through the teacher network improved performance when openness was lower than 64.4\%, but this contribution disappeared as openness increased (see Fig. \ref{fig_13}(b)). Introducing only the explorer had almost no affect; instead, it guided the student to perform worse. However, when the explorer and the teacher were adopted together, performance greatly increased. In addition, only when hints were provided by the teacher did the generated samples work well under conditions of high openness, as shown in Fig. \ref{fig_13}(c). The result reveals that the teacher and the explorer must be used together to create synergy.

\subsection{Comparison with State-of-the-Art Methods}
In this section, the proposed methods (T/E/S-CD and T/E/S-CDU) are compared with other state-of-the-art methods. We considered two different experimental setups. In the first setup, the unknown detection performance, which considers only the classification between “known” and “unknown”, was measured in terms of the area under the receiver operating curve (AUROC). In the second setup, the OSR performance, which reflects closed set classification with unknown detection, was measured in terms of the macroaverage F1-score.

For unknown detection performance comparison, we followed the protocol defined in \cite{Neal2018} with four image datasets: MNIST, SVHN, CIFAR-10, and Tiny-ImageNet. The MNIST, SVHN, and CIFAR-10 datasets were randomly partitioned into six known classes and four unknown classes. In addition, the model was trained on four nonanimal classes from CIFAR-10, and 10 animal classes were randomly selected from the CIFAR-100 dataset and added as unknown samples during the testing phase. This task is referred to as CIFAR+10. Similarly, 50 unknown classes were randomly selected from CIFAR-100, and we refer to this task as CIFAR+50. Finally, 20 classes were randomly chosen from the Tiny-ImageNet dataset as known classes, and the remaining 180 classes were set as unknown. For all datasets used for unknown detection, a random class split was repeated five times, and the averaged AUROC was used for evaluation. The comparison results are as shown in Table \ref{table_2}. Since the AUROC is a calibration-free measure, T/E/S-CD and T/E/S-CDU perform equally. Thus, we report the performance for both methods as T/E/S-CD(U).

Overall, the comparison results show that two-stage methods provide better performance than one-stage methods. This is because the two-stage methods were designed to maximize the unknown detection performance in training, leaving closed set classification as a task that is easily addressed by conventional DNNs. Despite the inherent weakness of the one-stage approach in unknown detection, the proposed T/E/S-CD(U) provided a competitive level of performance. Specifically, T/E/S-CD(U) performed best in all experimental settings among the one-stage approaches, even outperforming MLOSR and C2AE.

Finally, the proposed T/E/S learning was validated by comparison with the state-of-the-art methods in terms of OSR performance. The OSR models were trained on all training samples of the MNIST or the CIFAR-10. However, in testing, we additionally used another unknown dataset with the test samples from the MNIST or the CIFAR-10 datasets. Samples from Omniglot, MNIST-Noise, and Noise were considered unknown when MNIST was used as the training dataset. Here, Noise is a set of synthesized images in which each pixel value was independently sampled from a uniform distribution on [0, 1]. When the CIFAR-10 dataset was used for training, samples from ImageNet and LSUN were used as unknown samples. The ImageNet and LSUN datasets were resized or cropped to make the unknown samples the same size as the known samples, following the protocol suggested in \cite{Yoshihashi2019}. The known to unknown ratio was set to 1:1 for all cases.

The comparison results are shown in Tables \ref{table_3} and \ref{table_4}. T/E/S-CD performed the best on average, providing the highest score for the two unknown datasets when CIFAR-10 was trained. When MNIST was the training dataset, the proposed T/E/S-CD achieved the best results on all given unknown datasets. Contrary to the results of the unknown detection experiments, the proposed method outperformed CGDL, which performed the best in unknown detection, as well as other state-of-the art OSR methods.

\section{Conclusion}
In this paper, we developed a T/E/S learning method for OSR based on our intuition that the overgeneralization problem of deep learning classifiers can be significantly reduced after exploring various possibilities of unknowns. We first extended traditional T/S learning to HE-KD, not only to soften the posterior probabilities of the teacher network for known classes but also to extract uncertainty. Here, the softened probabilities prevent an unknown sample from obtaining a high score, and uncertainty is used as a hint that guides the explorer to discover unknown-like examples. In addition, to generate unknown-like open set examples, we introduced a new objective and training procedure to a GAN. The developed explorer networks explore a wide range of unknown possibilities. The experimental results showed that each component proposed in this paper contributes to the improvement in OSR performance. As a result, the proposed T/E/S learning method overwhelmed current state-of-the-art methods in terms of OSR performance.

Discriminating known and unknown samples is considered a key element of intelligent self-learning systems \cite{Boult2019}. However, if an identified unknown sample cannot be learned by a given system, then that system cannot be called a self-learning system. Thus, the proposed T/E/S learning should be extended so that it can incorporate class-incremental learning, where incoming unknown samples are used to continually train new unknown classes. This will be considered one of our future research directions.

\ifCLASSOPTIONcompsoc
  \section*{Acknowledgments}
\else
  \section*{Acknowledgment}
\fi

This work was supported by the National Research Foundation of Korea (NRF) grant funded by the Korean government (MSIT) (NRF-2019R1A2B5B01070358).

\ifCLASSOPTIONcaptionsoff
  \newpage
\fi



%



\bibliographystyle{IEEEtran}
\bibliography{Transactions-Bibliography/IEEEabrv,Transactions-Bibliography/reference}\ 


%

\begin{IEEEbiography}[{\includegraphics[width=1in,height=1.25in,clip,keepaspectratio]{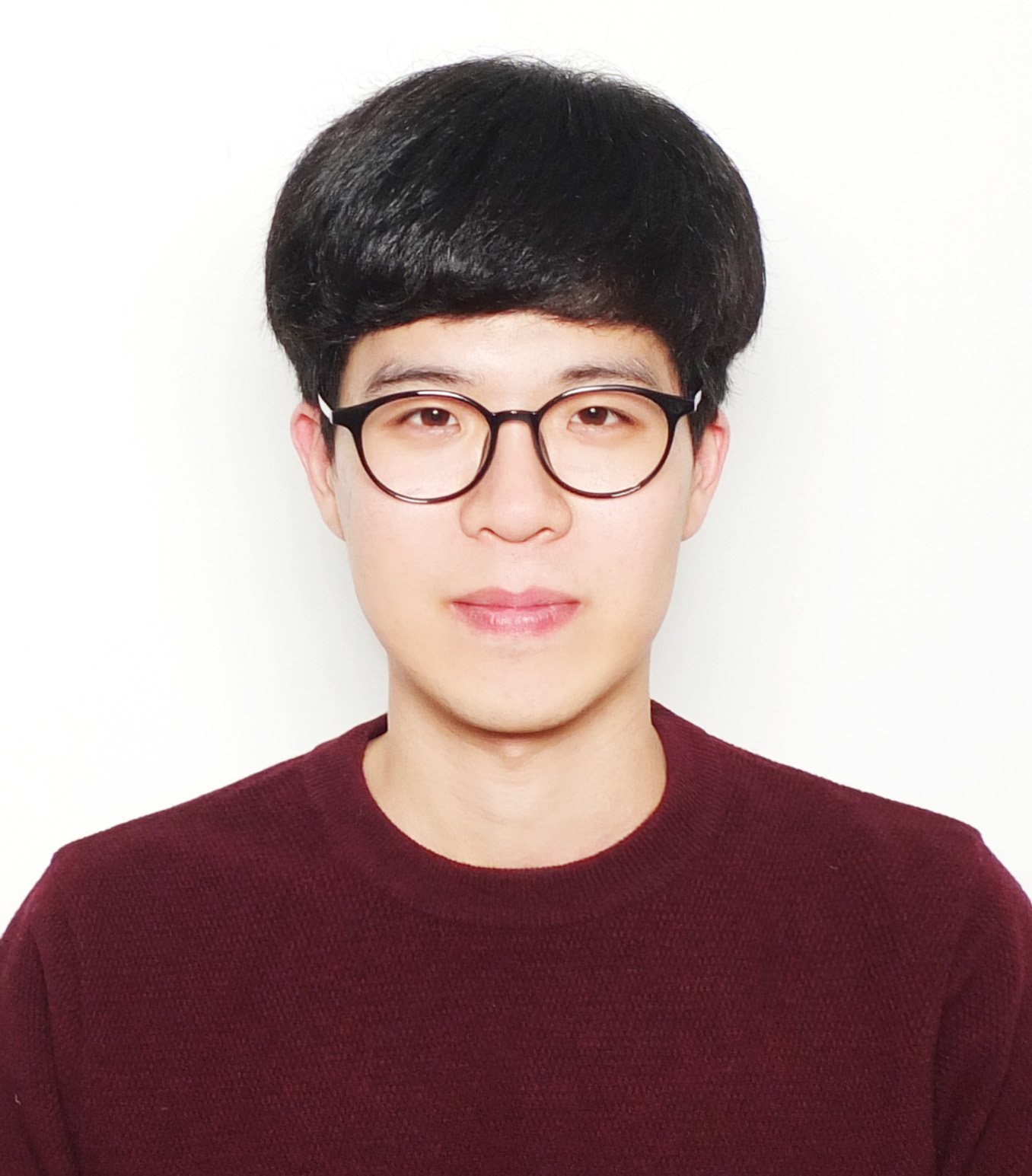}}]{Jaeyeon Jang}
received the Ph.D. degree in industrial engineering from Yonsei University, South Korea, in 2021, where he is currently a postdoctoral fellow. His current research interests include pattern recognition, machine learning, and reinforcement learning.
\end{IEEEbiography}

\begin{IEEEbiography}[{\includegraphics[width=1in,height=1.25in,clip,keepaspectratio]{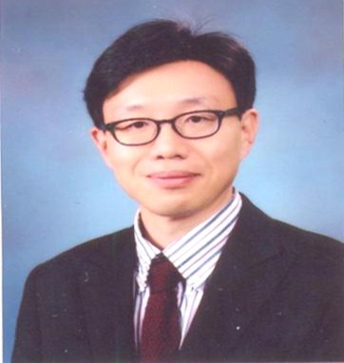}}]{Chang Ouk Kim}
received the Ph.D. degree in industrial engineering from Purdue University, West Lafayette, IN, USA, in 1996. He is currently a Professor with the Department of Industrial Engineering, Yonsei University, South Korea. He has published more than 100 papers in journals and conference proceedings. His current research interests include pattern recognition, machine learning, and data science for manufacturing and defense analysis.
\end{IEEEbiography}
\clearpage





\end{document}